\pdfoutput=1

\documentclass[11pt]{article}

\usepackage{ACL2023}

\usepackage{times}
\usepackage{latexsym}

\usepackage[T1]{fontenc}
\usepackage[utf8]{inputenc}
\usepackage{microtype}
\usepackage{inconsolata}
\usepackage{mathrsfs}
\usepackage{algorithm}  
\usepackage{algorithmicx}  
\usepackage{algpseudocode}
\usepackage{graphicx, subfigure}
\usepackage{makecell}
\usepackage{amsfonts,amssymb} 
\usepackage{amsmath}
\usepackage{multirow}
\usepackage{graphicx}
\usepackage{booktabs}

\title{Can Diffusion Model Achieve Better Performance in Text Generation? Bridging the Gap between Training and Inference!}

\author{Zecheng Tang$^{*}$, Pinzheng Wang\thanks{\; Equal contribution.}, Keyan Zhou, Juntao Li\thanks{\; Corresponding Author.}, Ziqiang Cao, Min Zhang \\  
 Institute of Computer Science and Technology, Soochow University, China \\
 \texttt{\{zctang,pzwang,kyzhou123\}@stu.suda.edu.cn}; \\
  \texttt{\{ljt,zqcao,minzhang\}@suda.edu.cn}
 }

\begin{document}
\maketitle
\begin{abstract}
Diffusion models have been successfully adapted to text generation tasks by mapping the discrete text into the continuous space.
However, there exist nonnegligible gaps between training and inference, owing to the absence of the forward process during inference.
Thus, the model only predicts based on the previously generated reverse noise rather than the noise computed by the forward process.
Besides, the widely-used downsampling strategy in speeding up the inference will cause the mismatch of diffusion trajectories between training and inference.
To understand and mitigate the above two types of training-inference 
 discrepancies, we launch a thorough preliminary study.
Based on our observations, we propose two simple yet effective methods to bridge the gaps mentioned above, named Distance Penalty and Adaptive Decay Sampling.
Extensive experiments on \textbf{6} generation tasks confirm the superiority of our methods, which can achieve $\mathbf{100}\times \rightarrow \mathbf{200}\times$ speedup with better performance. Our code is available at~\url{https://github.com/CODINNLG/Bridge_Gap_Diffusion}.

\end{abstract}

\section{Introduction}
With the prevalence of AIGC (Artificial Intelligence Generated Content) in recent years, generative models~\cite{kingma2013auto,goodfellow2020generative} have been receiving more attention.
As one of the representative generative models, diffusion models~\cite{sohl2015deep,song2020denoising} have achieved great success on myriads of generation tasks with continuous data, such as image ~\cite{song2020denoising,ramesh2022hierarchical,rombach2022high}, audio generation~\cite{kong2020diffwave}, and molecule generation~\cite{hoogeboom2022equivariant}, by iteratively refining the input noise to match a data distribution.
More recently, diffusion models have been successfully adapted to text generation~\cite{li2022diffusion, gong2022diffuseq,lin2022genie} by first leveraging an extra embedding module that maps the discrete data into the continuous space and then recovering the text from the continuous space with rounding strategy~\cite{li2022diffusion} or logits projection~\cite{strudel2022self}.

\begin{figure}[t]
    \centering
    \includegraphics[width=\columnwidth]{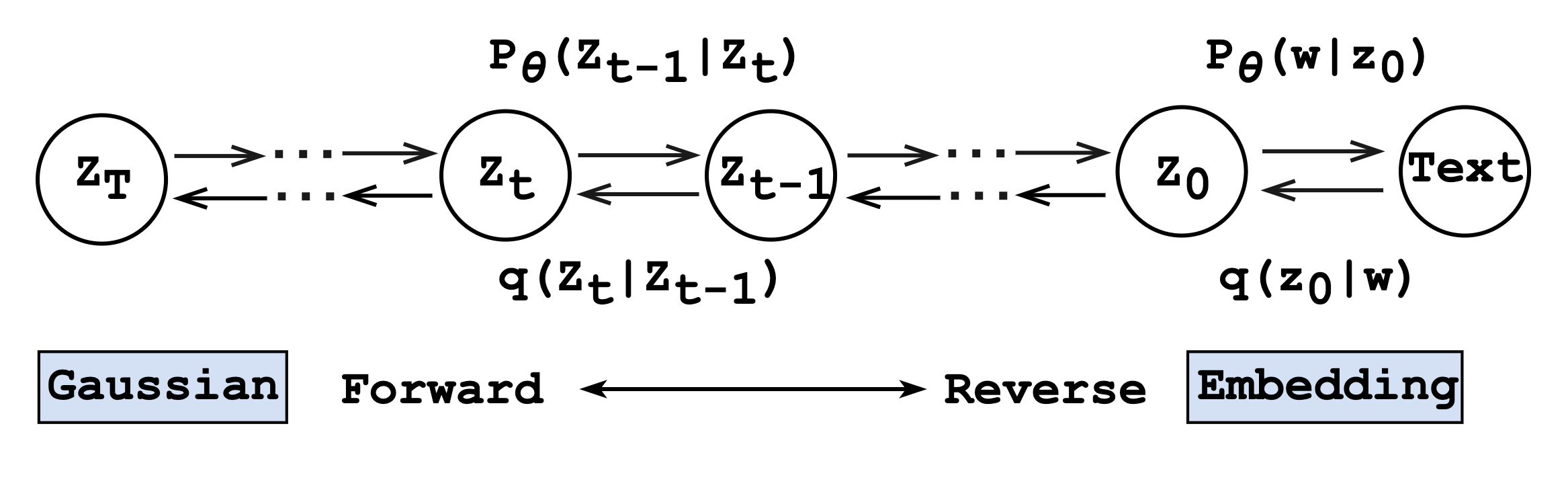}
    \caption{Overview of diffusion model for text generation, where $z_{t}$ denotes the intermediate noise at step $t$.}
    \label{fig:base_diffusion}
\end{figure}

A typical diffusion-based text generation model contains one reverse process~(from noise to data) and one forward process~(from data to noise), which is shown in Figure~\ref{fig:base_diffusion}.
More concretely, both of the two processes can be viewed as Markov chains, where the forward process gradually perturbs the data into Gaussian Noise while the reverse process recovers the original data step by step conditioned on the correlated noise from the forward process.
The training stage involves both of the above two processes, while the inference stage only consists of the reverse process, i.e., the model predicts based on the previous noise outputted by the model itself rather than the correlated forward noise.
Such discrepancy between training and inference, also called exposure bias~\cite{ranzato2015sequence}, leads to error accumulation as the denoising steps grow during the inference stage~\cite{huszar2015not,wiseman2016sequence}.

Another drawback of the diffusion model is that it requires multiple iterative denoising steps to produce the final results since the reverse process should approximate the forward process~\cite{ho2020denoising}, which usually involves thousands of steps.
Numerous iterative reverse steps of diffusion models are inevitably time-consuming for text generation. 
For instance, a diffusion model takes around 12 hours on one single NVIDIA A100 GPU to finish the inference of 10K sentences with a length of 128 while the CMLM-based non-autoregressive model~\cite{ghazvininejad2019mask} only takes a few minutes\footnote{Both diffusion model and CMLM model share the same backbone model, i.e., Transformer~\cite{vaswani2017attention}.}.
To accelerate the inference speed in text generation, down sampling~\cite{nichol2021improved} is leveraged~\cite{li2022diffusion, gao2022difformer, gong2022diffuseq}, though much faster but at the cost of performance owing to the gap between the downsampled steps in inference and the full diffusion trajectory in the training stage.

To explore the insights and the potential improvement of the aforementioned training-inference gaps, we conduct a preliminary study with a diffusion model~\cite{gong2022diffuseq} on the story generation task and mainly observe that: (1) injecting the noise generated by the model itself into the training stage can improve the model performance, and (2) the uniform downsampling strategy in the inference that treats each step equally impairs the model performance, and adaptive sampling strategy should be applied for different generation stages.
Accordingly, we propose two simple yet effective strategies: Distance Penalty and Adaptive Decay Sampling, to bridge the training-inference gaps and accelerate the inference process.
Experiments on \textbf{6} generation tasks of \textbf{3} different settings~(directed, open-ended, and controllable) show the superiority of our methods without changing the original architecture of the diffusion model or adding more parameters.
Surprisingly, our methods can achieve $\mathbf{100}\times$ speedup with performance improvement or $\mathbf{200}\times$ acceleration with competitive results.

\section{Background}

\subsection{Diffusion Model}
Diffusion models are one of the prevalent generative models~\cite{sohl2015deep,song2020denoising,nichol2021improved}, which can transfer an arbitrary data distribution into the Gaussian noise with the forward process and recover the data from the pure noise with the reverse process and both two processes can be regarded as a Markov chain.
Specifically, given the time steps $\mathcal{T}=\{0, 1, \cdots, T\}$ and the original data distribution $z_{0}$ at time step $t=0$, the forward process gradually perturbs it into the Gaussian noise $z_{T}\sim \mathcal{N}(0, \mathbf{I})$ at time step $t=T$:
\begin{equation}
    \label{equ:q}
    q(z_{t}\mid z_{t-1}) = \mathcal{N}(z_{t};\sqrt{1-\beta_{t}}z_{t-1},\beta_{t}\mathbf{I}),
\end{equation}
where $z_{t}$ represents the intermediate noise at time step $t$ and $\beta_{t}\in (0, 1)$ is the scaling factor, controlling the amount of added noise at time step $t$.

The reverse diffusion process recovers the initial data distribution $z_{0}$ from the Gaussian noise $z_{T}$ by predicting the noise of current time step $t$ and denoising it into the next reverse state $z_{t-1}$:
\begin{equation}
    \label{equ:p_theta}
    p_{\theta}(z_{t-1}\mid z_{t}) = \mathcal{N}(z_{t-1}; \mu_{\theta}(z_{t}, t), \Sigma_{\theta}(z_{t}, t)),
\end{equation}
where $\mu_{\theta}$ and $\Sigma_{\theta}$ can be implemented by neural networks $f_{\theta}$, e.g., Transformer\footnote{$\Sigma_{\theta}$ is often set as $\sigma^{2}_{t}\mathrm{I}$~\cite{ho2020denoising}, where $\sigma^{2}_{t}=\beta_{t}$.}:
\begin{equation}
\label{equ:mu_function}
    \mu_{\theta}(z_{t}, t) = \frac{1}{\sqrt{\alpha_{t}}}(z_{t} - \frac{\beta_{t}}{\sqrt{1 - \bar{\alpha}_{t}}}f_{\theta}(z_{t}, t)),
\end{equation}
where $\alpha_{t} = 1 - \beta_{t}$ and $\bar{\alpha}_{t} = \prod_{i=1}^{t}\alpha_{i}$.

\paragraph{Training}
The training objective of the diffusion model is to maximize the marginal likelihood of data $\log p_{\theta}(z_{0})$, and the simplified training objective can be written as~\cite{ho2020denoising}:
\begin{equation}
\small
\label{equ:simple_objective}
\mathcal{L}_{simple} = \sum\limits_{t=1}^{T}\mathop{\mathbb{E}}\limits_{q(z_{t}\mid z_{0})} || \mu_{\theta}(z_{t}, t) - \hat{\mu}({z_{t}, z_{0}})||^{2},
\end{equation}
where $\hat{\mu}(z_{t},z_{0})$ is the mean of $q(z_{t-1}\mid z_{0}, z_{t})$, and it is worth noting that each intermediate noise $z_{t}$ can be obtained directly without the previous history during the training stage~(Equation~\ref{equ:q_extend}).

\paragraph{Inference}
The inference stage only consists of the reverse process. 
To sample $z_{t-1}\sim p_{\theta}(z_{t-1}\mid z_{t})$ in Equation~\ref{equ:p_theta}, reparameterization strategy~\cite{kingma2013auto} is leveraged:
\begin{equation}
    \label{equ:sampling}
    z_{t-1} = \mu_{\theta}(z_{t}, t) + \sigma_{t}\epsilon,
\end{equation}
where $\epsilon\sim \mathcal{N}(0, \mathbf{I})$, $\sigma_{t}^{2} = \beta_{t}$, and $z_{t}$ is initialized with pure Gaussian noise in the beginning.
More details about the training and inference stages as well as the derivations are shown in Appendix~\ref{appendix:detivation_train_object}.

\subsection{Diffusion Model for Text Generation}
\label{sec:diffsion_for_text_generation}
The core of applying diffusion models for text generation task is the transition between discrete space and continuous space.
Existing works mainly introduce the embedding function~\cite{li2022diffusion} $\mathcal{E}(\cdot)$ to map the discrete text $\mathbf{w} = \{w_1, w_2, \cdots, w_L\}$ of length $L$ into the continuous space $\mathcal{E}(\mathbf{w})=\{\mathcal{E}(w_1),\mathcal{E}(w_2),\cdots,\mathcal{E}(w_L)\}\in \mathbb{R}^{Ld}$.
Thus, the diffusion model can handle discrete text generation by adding an extra forward step before $t=0$, denoted as $q(z_{0}\mid \mathbf{w})=\mathcal{N}(\mathcal{E}(\mathbf{w}), \sigma_{0}\mathbf{I})$, and another step at the end of the reverse process, i.e., $p_{\theta}(\mathbf{w}\mid z_{0})$.
More details are given in Appendix~\ref{appendix:embedding}.

\subsection{Inference Speedup}
One critical point that prevents the usability of diffusion models in text generation is their slow sampling speed during inference due to the long reverse trajectory, which makes each diffusion step simple and easy to estimate~\cite{sohl2015deep}.
To accelerate the inference speed in text generation tasks, current works~\cite{li2022diffusion, gao2022difformer} apply the downsampling strategy~\cite{nichol2021improved} that picks the subset $\mathcal{T}^{\prime}=\{t_{1}^{\prime}, t_{2}^{\prime}, \cdots, t_{k}^{\prime}\}$ from the full diffusion trajectory and each intermediate reverse step can be obtained by: $z_{t-1}^{\prime} = \mu_{\theta}(z_{t}^{\prime}, t^{\prime}) + \sigma_{t}^{\prime}\epsilon$.


\section{Gaps between Training and Inference}
From the above description of diffusion models, we can summarize two gaps: (1) \textit{the reverse process at time step $t$ in inference is conditioned on the predicted noise $z_{t+1}$ by the model itself while $z_{t+1}$ can be obtained directly with the forward computation $q(z_{t+1}\mid z_0)$ during training}, and (2) \textit{the downsampled time subset $\mathcal{T}^{\prime}$ in inference is inconsistent with the full diffusion trajectory $\mathcal{T}$ in training stage when applying the downsampling method for inference speedup.} 
To calibrate the effects of these two types of training-inference gaps, we launch a study on the story generation task in this section.

\subsection{Study Settings}
We implement the diffusion model with the transformer model and select the ROC Stories (ROC) corpus~\cite{mostafazadeh2016corpus} for the story generation task.
Specifically, given the prompt or the source sentence $\mathbf{w}^{x}$ and the reference $\mathbf{w}^{y}$, we apply the partially noising strategy~\cite{gong2022diffuseq} for training~(Appendix~\ref{appendix:detivation_train_object}).
We utilize BLEU (B-2) score~\cite{papineni2002bleu} to reflect the generation precision (the higher, the better), Lexical Repetition (LR-2) score~\cite{shao2019long} to show the diversity of text (the lower, the better), ROUGE (R-2) to represent the recall of generation result (the higher, the better) and Perplexity (PPL) to reflects the fluency (the lower, the better).
More implementation details are in Appendix~\ref{appendix:implementation_details}.

\subsection{Analysis}
\label{sec:preliminary_analysis}

\begin{figure}[t] 
\centering
\subfigure[B-2 scores.] { 
    \label{fig:noise_exp_bleu2}
    \includegraphics[width=0.3\columnwidth]{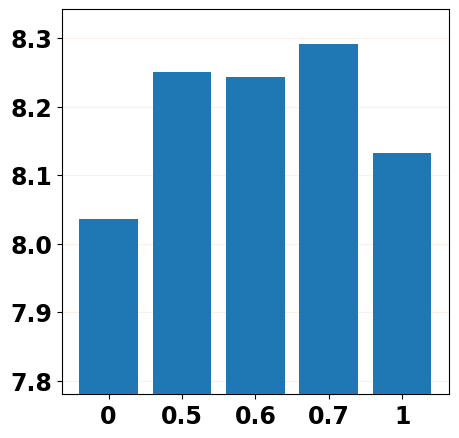}
}
\subfigure[LR-2 scores.] { 
    \label{fig:noise_exp_lr}
    \includegraphics[width=0.3\columnwidth]{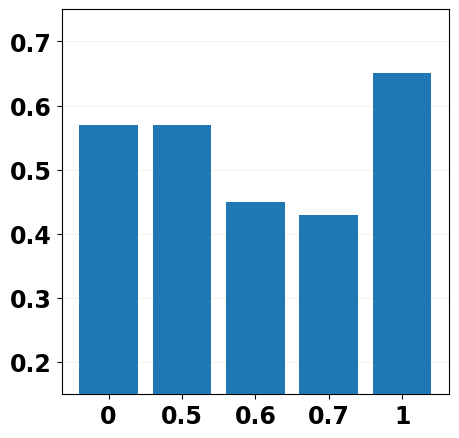}
}
\subfigure[PPL scores.] { 
    \label{fig:noise_exp_lr}
    \includegraphics[width=0.3\columnwidth]{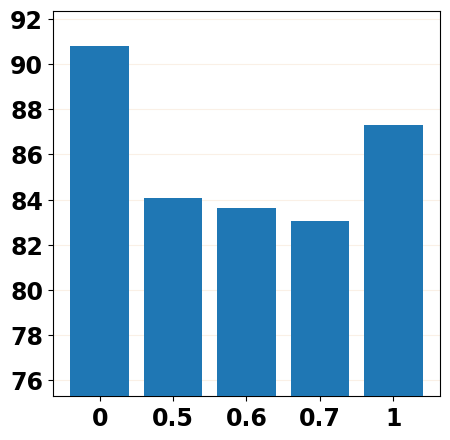}
}
\caption{Evaluation results of noise injection, where the number in abscissa represents $\gamma_2$ and $\gamma_1=1-\gamma_2$.}
\label{fig:preliminary_noise_exposure}
\end{figure}

\paragraph{Training with Predicted Noise}
To mitigate the training-inference gap, it is natural to inject part of the predicted noises into the training stage by replacing the forward noise $z_{t+1}$ in $p_{\theta}(z_{t}\mid z_{t+1})$ with the predicted noise $z_{t+1}^{\prime}$ from the $(t+1)$-th step of the reverse process or injecting the predicted noise into $z_{t}$ by replacing $||\mu_{\theta}(z_{t}, t) - \hat{\mu}(z_{t}, z_{0})||^{2}$ in Equation~\ref{equ:simple_objective} with $\gamma_{1}||\mu_{\theta}(z_{t}, t) - \hat{\mu}(z_{t}, z_{0})||^{2} + \gamma_{2}||\mu_{\theta}(z_{t}, t) - \hat{\mu}(z_{t}^{\prime}, t)||^{2}$, where $z_{t}\sim q(z_{t}\mid z_{0})$ and $z_{t}^{\prime}\sim p_{\theta}(z_{t}\mid z_{t+1}^{\prime})$.
We report the evaluation results in Figure~\ref{fig:preliminary_noise_exposure} with different settings of $\gamma_1$ and $\gamma_2$ and can mainly observe that replacing the forward noise with the predicted noise~($\gamma_2=1, \gamma_1=0$) does mitigate the training-inference gap by achieving a better performance than the vanilla training scheme ($\gamma_2=0, \gamma_1=1$), and the injecting strategy performs better than the replacing one.
More details about noise replacement operation and evaluation results are shown in Appendix~\ref{appendix:noise_injection}.

\begin{figure}[t] 
\centering
\subfigure[B-2 of non-uniform sampling.] {
    \label{fig:non_uniform_sampling_bleu2}
    \includegraphics[width=0.98\columnwidth]{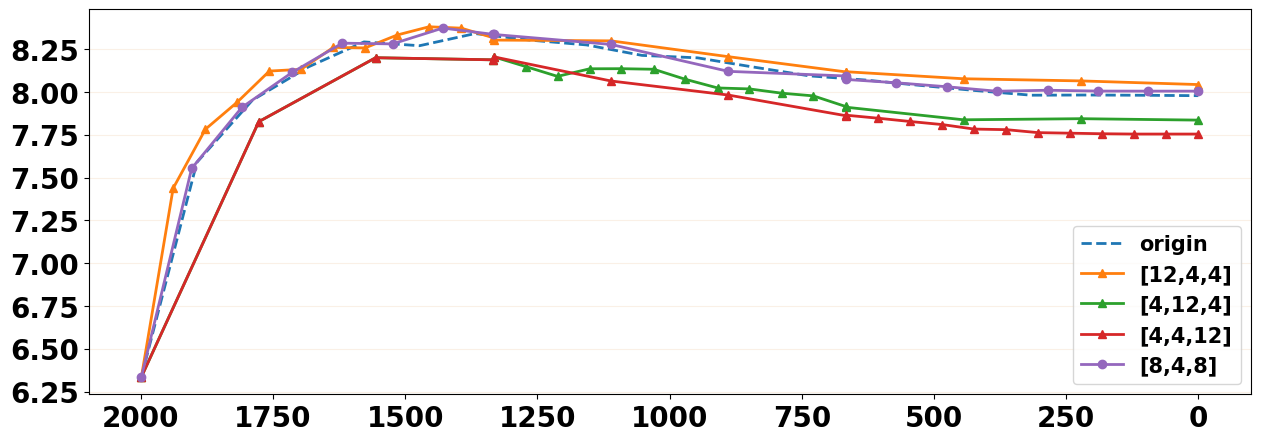}
}
\subfigure[R-2 of non-uniform sampling.] { 
    \label{fig:non_uniform_sampling_ppl}
    \includegraphics[width=0.98\columnwidth]{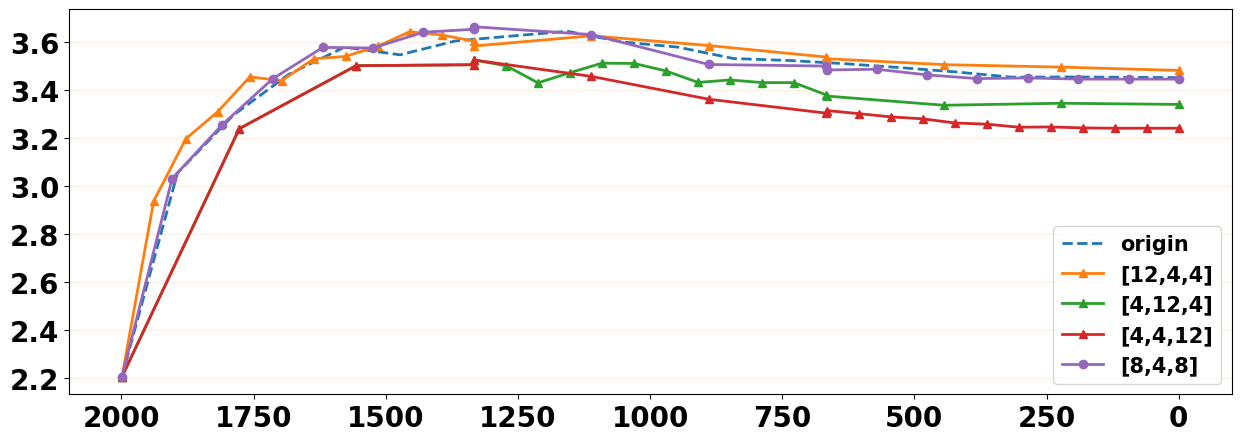}
}
\caption{Comparison between non-uniform steps of $[\kappa_1, \kappa_2, \kappa_3]$ and the original uniform scheme, where 
the x-axis represents the denoising steps, and y-axis  illustrates the evaluation results for each metric. The dots on each curve indicate the number of down-sampled steps.}
\label{fig:preliminary_noise_non_uniform}
\end{figure}

\paragraph{Sampling Strategy}
\label{sec:sampling_strategy_preliminary}
Downsampling can accelerate the inference by uniformly selecting the subsets $\mathcal{T}^{\prime}$ from the full diffusion trajectory $\mathcal{T}$ but at the cost of performance.
Such a uniform sampling strategy treats each reverse step equally while neglecting the discrepancies among them in contribution to the final result.
To explore whether such an equal-step sampling strategy brings the performance decrease, we simply compare different non-uniform sampling schemes.
Along with the reverse steps, we split the reverse process into three stages $[\kappa_1, \kappa_2, \kappa_3]$ and downsample different numbers of steps for each stage but keep the total downsampled steps the same\footnote{For total number of downsampled steps 20, we can sample $\{[12, 4, 4], [4,12,4], [4,4,12], [8,4,8]\}$ steps as $[\kappa_1, \kappa_2, \kappa_3]$.}.
As shown in Figure~\ref{fig:preliminary_noise_non_uniform}, we can observe that when downsampling more steps from $\kappa_{1}$~(\textcolor{orange}{orange curve}), the model can achieve a better performance than other downsampling schemes~(\textcolor[RGB]{66,170,66}{green curve}, \textcolor{red}{red curve}, and \textcolor[RGB]{148,103,189}{purple curve}) and even exceed the original full reverse steps~(\textcolor{blue}{blue curve}).
In other words, the equal-step uniform downsampling scheme does limit the model capability, and the simple non-uniform downsampling strategy can mitigate such issue and meanwhile accelerate the inference speed.

\paragraph{Extensive Trials}
As mentioned above, the gap brought by the different diffusion trajectories in the inference stage, i.e., downsampled reverse steps v.s. the full reverse steps, further aggravates the training-inference discrepancy.
In view that simply injecting the predicted reverse noise in training can effectively narrow the gaps between training and inference, it is also appealing to make such a strategy adapt to the downsampled diffusion trajectories, i.e., introducing the downsampled reverse noises in the training stage.
For instance, we can inject the predicted reverse noise downsampled from the reverse steps of $(t,t+\delta]$ into the $d$-th~($d\sim (t,t+\delta]$) forward noise to compute the $t$-th step reverse noise, i.e., replacing the forward noise $z_{t+1}$ in $p_{\theta}(z_{t}\mid z_{t+1})$ with $z_{d\sim (t,t+\delta]}$.

\begin{figure}[t] 
    \centering
    \includegraphics[width=0.9\columnwidth]{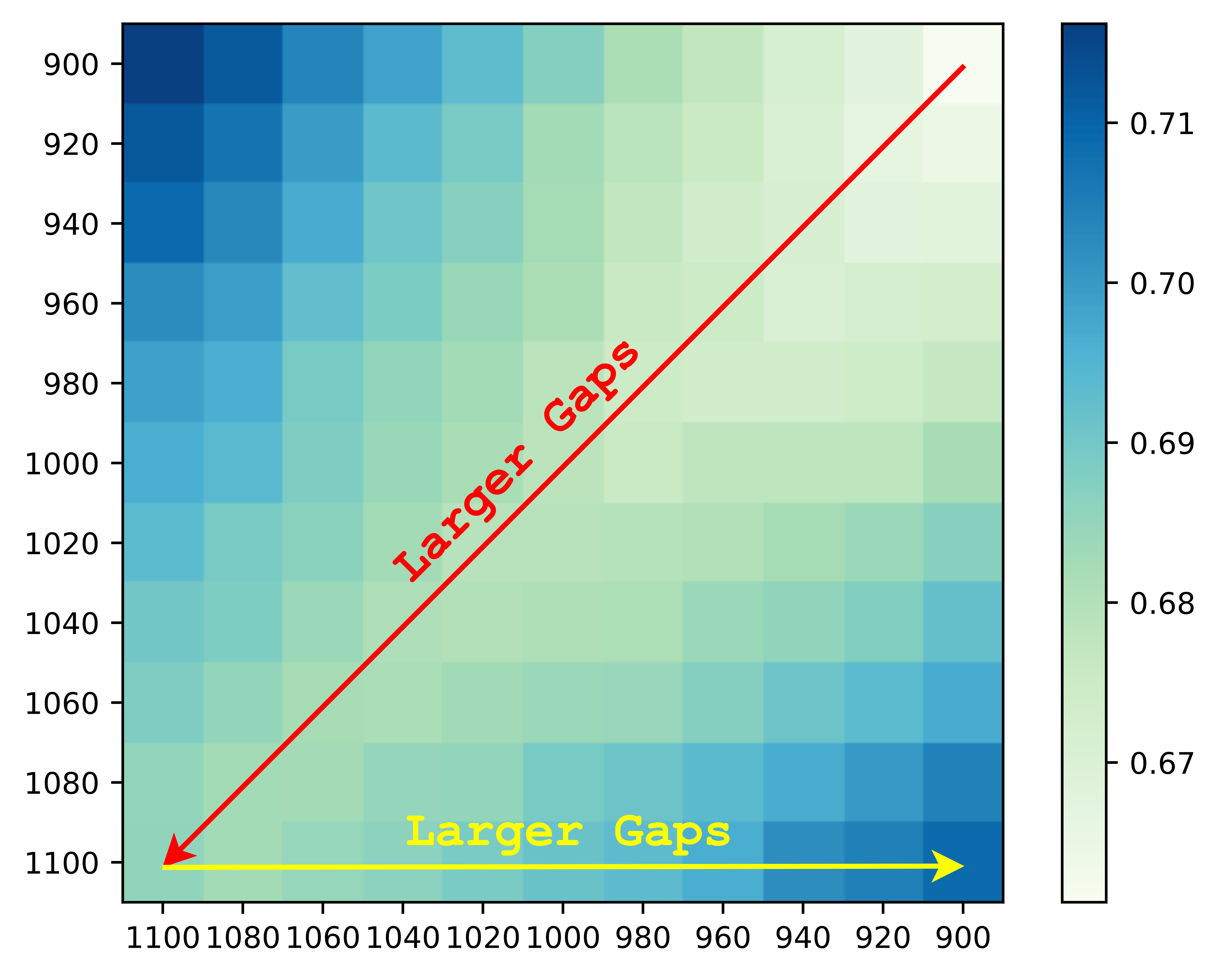}
    \caption{Euclidean distance between the predicted reverse noise and the forward noise of a converged model.}
    \label{fig:diff_dis}
\end{figure}

Intuitively, adding a perturbation with a reasonable range of values in training can make the model more robust towards the perturbation during inference, while an unconstrained perturbation value might risk the model training, e.g., the training collapse in auto-regressive text generation models~\cite{zhang2019bridging}.
For our purposes, the discrepancy before and after injecting the downsampled reverse noise in each training step should fall in a rational range, which mainly depends on the time step $t$ and the choice of $\delta$.
To explore more insights, we depict the discrepancy between predicted reverse noises and forward noises along with 200 randomly selected continuous time steps with the Euclidean distance, which is consistent with the training objective in Equation~\ref{equ:simple_objective}.
To simplify the study experiment, we downsample a time step for every twenty steps\footnote{We utilize the diffusion model trained with 240K steps. More implementation details are shown in Appendix~\ref{appendix:pseudo_target_selection}}.
As shown in Figure~\ref{fig:diff_dis}, we can observe that (1) the discrepancy between predicted reverse noises and forward noises is getting larger along with the increase of time step $t$ (red diagonal arrow), and (2) the differences between the forward noise at time step $t$ and the predicted reverse noise from $t$ to $t+\delta$ are becoming larger along with the increase of time step (yellow horizontal arrow).
Thus, the range of downsampled reverse noise steps should be gradually narrowed along with the increase of time step.

\subsection{Potential Improvement}
Based on the analysis mentioned above, we can conclude that: \textbf{(1)} injecting the predicted reverse noise into the training stage can mitigate the training-inference gaps, \textbf{(2)} the scheme of uniform downsampling in inference which treats each step equally harms the model performance, and a non-uniform adaptive method should be designed, and \textbf{(3)} inspired by (1) and (2), we can inject the downsampled reverse noises into the training stage while the range of downsampled steps should be gradually narrowed as the time step increases.

\begin{figure*}[t]
    \centering
    \includegraphics[width=1.0\textwidth]{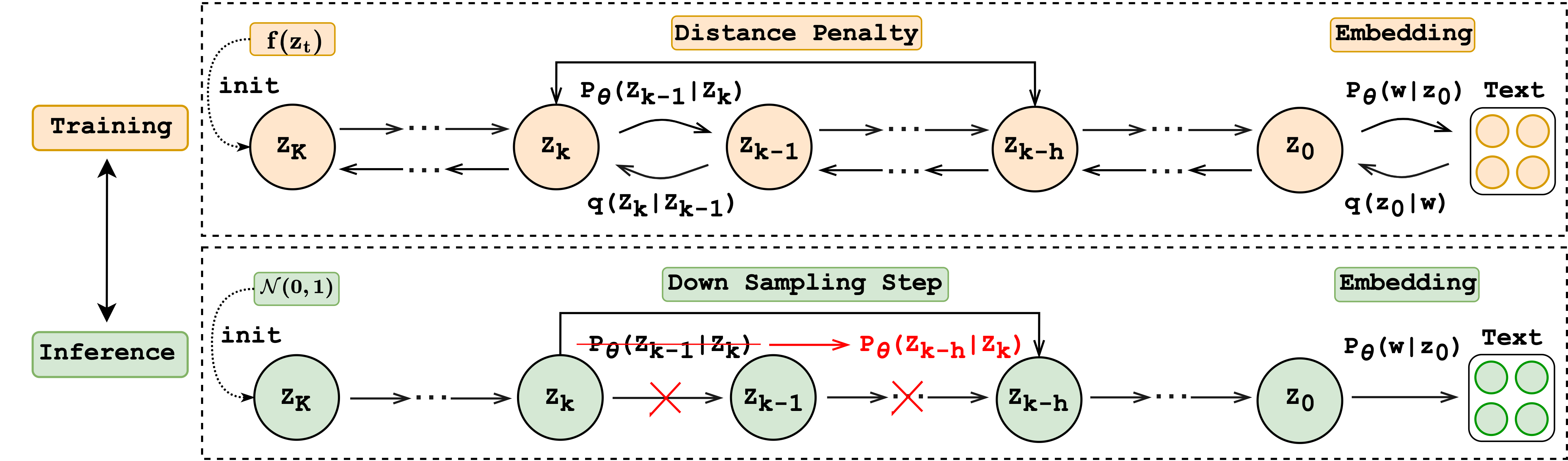}
    \caption{Overview of Distance Penalty method for bridging the gaps between training and inference.}
    \label{fig:dispen_diffusion}
\end{figure*}

\section{Method}
We propose two simple yet effective methods: \textbf{Distance Penalty} in the post-training stage and \textbf{Adaptive Sparse Sampling} in inference to bridge the gaps without introducing any architecture modification to diffusion models.
Thus, it can be flexibly adapted to different diffusion model variants.

\subsection{Distance Penalty}
We first introduce the Distance Penalty strategy, which injects the Downsampled predicted reverse noise into the post-training stage of diffusion models that consists of $T$ time steps.\footnote{To make sure each reverse step can generate a rational noise for perturbation, we apply the Distance Penalty strategy to a converged model. We also apply such a strategy in different training stages in Appendix~\ref{appendix:distance_penalty}.}
For better illustration, we utilize new symbols $\mathcal{K}=\{0, 1, \cdots, K\}$ for the time steps in the post-training stage to distinguish from the original diffusion trajectory $\mathcal{T}$ in the training stage.
The overview of the Distance Penalty strategy is shown in Figure~\ref{fig:dispen_diffusion}.

\paragraph{Downsampling Range in Training}
To obtain a rational predicted reverse noise for each step $k$, i.e., conduct the downsampling operation in the range $\mathcal{R}_{k}=\{k-1, \cdots, k-h\}$, and mitigate the training-inference gaps, we constrain the total amount of noises in $\mathcal{R}_{k}$ with the threshold $\omega^{k}_{adj}$:
\begin{equation}
    \label{equ:threshhold}
    \omega^{k}_{adj} = \frac{\sqrt{1 - \bar{\alpha}_{K}}}{k^{\prime}},
\end{equation}
where $\sqrt{1 - \bar{\alpha}_{K}}$ denotes the scaling factor that controls the variance of noise accumulated at step $K$~(Appendix~\ref{appendix:detivation_train_object}), and $k^{\prime}$ is the number of the predefined downsampled steps in inference.

\paragraph{Noise Injection}
After obtaining the downsampling range $\mathcal{R}_{k}=\{k-1, \cdots, k-h\}$ for step $k$, we can inject the predicted reverse noise into reverse step $k$ with Equation~\ref{equ:q_extend} in Appendix~\ref{appendix:detivation_train_object}, by which we can acquire every predicted reverse noise $z_{d\sim \mathcal{R}_{k}}$ with the correlated forward noise $z_{d+1}$:
\begin{equation}
\label{equ:original_distance_penalty}
\small
\left\{
\begin{aligned}
    & \mathcal{L}_{dis} = \sum\limits_{k=1}^{K}\sum\limits_{h=1}^{|\mathcal{R}_{k}|}||\mu_{\theta}(z_{k}, k) - \hat{\mu}(z_{k-h}, k-h)||^{2} \\
    & z_{k-h} = \mu_{\theta}(z_{k-h+1}, k-h+1) + \sigma^{2}_{k-h+1}\mathbf{I},
\end{aligned}
\right.
\end{equation}
where $\mathbf{I}\sim \mathcal{N}(0, 1)$ and $\sigma^{2}_{k-h+1}=\beta_{k-h+1}$.

However, the random item $\mathbf{I}$ in $z_{k-h}$ makes 
the model difficult to converge, and the computation of $\mathcal{L}_{dis}$ is complex.
Thus, we apply the simplified Equation~\ref{equ:train_objective_for_nlp} in Appendix~\ref{appendix:embedding}, which approximates the model output $f_{\theta}(z_d)$ to the original data distribution directly, and rewrite the loss function into:
\begin{equation}
\small
\mathcal{L}^{dis}_{simple} = \sum\limits_{k=1}^{K}\sum\limits_{h=1}^{|\mathcal{R}_{k}|}||f_{\theta}(z_{k}, k) - f_{\theta}(z_{k-h}, k-h)||^{2}, 
\end{equation}

Considering that the step $k$ is uniformly sampled during the training stage, to avoid the boundary condition $k-h<0$, the final training objective is:
\begin{equation}
    \mathcal{L}_{post} = \left\{
    \begin{aligned}
        & \mathcal{L}_{simple} + \gamma \mathcal{L}_{simple}^{dis} & k\geq h &\\
        & \mathcal{L}_{simple} & k < h &,
    \end{aligned}
    \right.
\end{equation}
where $\mathcal{L}_{simple}$ is the original training objective of diffusion model~(Equation~\ref{equ:simple_objective}), and $\gamma$ is the penalty weight that controls the degree of the constraint.

\subsection{Adaptive Decay Sampling}
\label{sec:ADS}
We also apply the Adaptive Decay Sampling~(ADS) strategy to mitigate the issues brought by uniform downsampling.
More concretely, we split the reverse process into three stages $[\kappa_1, \kappa_2, \kappa_3]$ and adaptively adjust the downsampled steps in each stage according to the total amount of added noise of each stage during the training, i.e., more downsampled steps are required to decompose the large noise, which is controlled by $\bar{\alpha}_{1:k}$~(Equation~\ref{equ:mu_function}):
\begin{equation}
\small
\eta_{i} = \left\{
\begin{aligned}
    & \frac{1}{\sqrt{1 - \bar{\alpha}_{Ki/3}}} - \sum\limits_{j=1}^{i-1}\eta_{j} & i > 1 & \\
    & \frac{1}{\sqrt{1 - \bar{\alpha}_{K/3}}} & i=1 &
\end{aligned}
     \right.
\end{equation}
Then, we can treat $\eta_{i}$ as the weight to split the total downsampled steps $\mathcal{T}^{\prime}$ into different subsets for each stage $\kappa_{i}$.
Such a strategy is associated with the noise scheduler, which controls the calculation of $\bar{\alpha}_{1:k}$, and we put more details in Appendix~\ref{appendix:adaptive_decay_sampling}.
\section{Experiment}

\subsection{Settings}
We describe the main settings of our experiments in this section, and more implementation details can be referred to Appendix~\ref{appendix:implementation_details}.
\paragraph{Tasks \& Datasets}
We conduct the experiments on three different generation tasks, i.e., directed, open-ended, and controllable.
For directed generation tasks, we utilize the WIkI-AUTO corpus~\cite{jiang2020neural} for Text Simplification task and Quora Question Pairs\footnote{\url{https://www.kaggle.com/c/quora-question-pairs}} corpus for Paraphrase task.
For open-ended generation tasks, we adopt the ROC Stories~(ROC) corpus for Story Generation task and Quasar-T~\cite{dhingra2017quasar} dataset preprocessed by \citet{lin2018denoising} and \citet{gong2022diffuseq}\footnote{\url{https://drive.google.com/drive/folders/122YK0IElSnGZbPMigXrduTVL1geB4wEW?usp=sharing}} for Question Generation task.
For controllable text generation task, we utilize the E2E~\cite{novikova2017e2e} dataset and select Semantic Content control task and Syntax Spans control task. More statistics of datasets are listed in Table~\ref{tab:statistic_dataset} of Appendix~\ref{appendix:implementation_details_datasets}.

\paragraph{Baselines}
We apply the DIFFSEQ~\cite{gong2022diffuseq} as the baseline for directed generation tasks and open-ended generation tasks and utilize Diffusion-LM~\cite{li2022diffusion} for controllable generation tasks.
For both of the above two baselines, we implement $f_{\theta}$ with Transformer model and select the converged checkpoint for post-training.
The total diffusion steps $T$ for training and $K$ for post-training are both 2,000.
We set the hyper-parameter $\gamma$ of $\mathcal{L}_{simple}^{dis}$ as 2 and utilize the square-root noise schedule for $\beta_{t}$.
Besides, we also compare the generation results with the autoregressive~(AR) model BART~\cite{lewis2020bart} and the none-autoregressive~(NAR) model CMLM~\cite{ghazvininejad2019mask}, which are both implemented with the open library \textit{Fairseq} toolkit\footnote{\url{https://github.com/facebookresearch/fairseq}}~\cite{ott2019fairseq}.
For open-ended generation task, we utilize nucleus sampling~\cite{holtzman2019curious}~(top-$p$=0.5) for the BART model.
Meanwhile, for the controllable generation tasks, we apply PPLM~\cite{dathathri2019plug} and FUDGE~\cite{yang2021fudge} for guidance.

\paragraph{Evaluation Metrics}
For open-ended generation tasks, we report BLEU~(B-$n$)~\cite{papineni2002bleu}, ROUGE~(R-n)~\cite{lin2004rouge}, Distinct (D-$n$)~\cite{li2016diversity}, Lexical Repetition (Rep-$n$, 4-gram repetition for $n$-times)~\cite{shao2019long}, BERTScore~\cite{zhang2019bertscore}, Mauve score~(Mav)~\cite{pillutla2021mauve}, Perplexity~(PPL), and Semantic Similarity (SIM, semantic similarity between generations and corresponding prompts)~\cite{guan2021long}\footnote{\url{https://huggingface.co/sentence-transformers/bert-base-nli-mean-tokens}}.
We select part of the metrics mentioned above for the evaluation of directed generation tasks and utilize success rate~(Ctrl)~\cite{li2022diffusion} to evaluate the control effect.
The setting of $n$ is described in each subsection below, and the evaluation results~(D-$n$, Rep-$n$, and PPL) of the golden text are reported for comparison.
For fair comparison, we calculate the PPL and Mauve score with the same pre-trained GPT-2~\cite{radford2019language} model\footnote{\url{https://huggingface.co/gpt2-large}}.
We also fine-tune the GPT-2 model on each downstream dataset and report their PPL and Mauve score in Appendix~\ref{appendix:eval_fine_tuned_gpt2}.

\begin{table*}[t]
    \centering
    \small
    \resizebox{\textwidth}{!}{
    \begin{tabular}{l| l l l l l l l l l c l l}
    \toprule
    \textbf{Data} & \textbf{Model} & \textbf{Step} & \textbf{B-2$(\uparrow$)} & \textbf{B-4$(\uparrow$)} & \textbf{R-2$(\uparrow$)} & \textbf{R-L$(\uparrow$)} & \textbf{D-2$(\uparrow$)} &  \textbf{LR-2$(\downarrow$)} & \textbf{BS$(\uparrow$)} & \textbf{Mav$(\uparrow$)} & \textbf{$\Delta$PPL$(\downarrow$)} & \textbf{ SIM$(\downarrow$)}\\
    \midrule
    \multirow{7}{*}{ROC} 
    & CMLM & - & 8.17 & 2.52 & \underline{4.36} & 19.74 & 14.95 &  25.60 & 51.68 & 2.73  & 13.45~(+)  & \underline{15.38}\\ 
    & BART & - & 7.55 & 2.38 & 3.95 & 18.83 & 15.88 &  0.98 & \underline{57.01} & \underline{70.64}  & 2.95~(-)  & 16.19 \\
    \cmidrule{2-13}      
    &  DIFFSEQ$\dagger$ & 2,000 & 8.39 & 2.48 & 3.81 & 18.88 & \bf 22.64 & \bf 0.71 & 54.19 & 34.45  & 49.44~(+) & 16.03\\
    &  + Ours$\dagger$ & 2,000 & \bf 8.90 & \bf 2.66 & \bf 4.27 & \bf 19.59 & 19.86 &  1.22 & \bf 54.91 & \bf 41.56  & \textbf{33.56}~(+)  & \bf 15.94 \\
    \cmidrule{2-13}  
    &  Respace$\ddagger$ & 20 & 8.43 & 2.48 & 3.83 & 18.87 & \bf 22.93 & \bf 0.75 &  53.80 & 25.37  & 56.32~(+) & 16.06\\
    &   + Ours$\ddagger$ & 20 & \bf 8.86 & \bf 2.63 & \bf 4.16 & \bf 19.58 & 21.48 &  0.90 & \bf 54.05 & \bf 28.06  &\textbf{48.76}~(+)  & \bf 15.96 \\
     \cmidrule{2-13}  
    &  Golden & - & - &- &- &- & 36.50 & 0.02 &- & -& 29.72  & 16.49 \\
    
    \midrule
    \multirow{7}{*}{Quasar-T} 
    &  CMLM & - & 13.37 & 7.69 & 12.19 & 26.26 & 9.95 & 15.70  & 50.53 & 1.96 & 88.37~(+)  & 17.38\\ 
    & BART & - & 11.92 & 7.45 & 11.07 & 23.34 & 10.87  & \underline{0.00} & 57.07 & 3.09  &  0.75~(+)  & 18.08\\
    \cmidrule{2-13}      
    &  DIFFSEQ$\dagger$ & 2,000 & 23.50 & 17.11 & 23.10 &  36.32 & \bf 21.95 &  \bf 11.34 & 62.25 & 4.68  & 95.68~(+)  & \bf 15.84\\
    &  + Ours$\dagger$ & 2,000 & \bf 23.67 & \bf 17.53 & \bf 23.34 & \bf 36.55 & 19.75  & 12.07 & \bf 62.80 & \bf 10.91  &  \textbf{58.68}~(+)  &  15.87\\
    \cmidrule{2-13}  
    &  Respace$\ddagger$ & 20 & 23.15 & 16.92 & 22.75 & 35.97 & \bf 25.20 & \bf 10.31 & 61.76 & 4.53  &  169.75~(+) & \bf 15.80\\
    &   + Ours$\ddagger$ & 20 & \bf 23.55 & \bf 17.45 & \bf 23.17 & \bf 36.02 & 21.18  & 11.23 & \bf 62.52 & \bf 5.41 & \textbf{96.83}~(+)  & 15.85\\
     \cmidrule{2-13}  
    &  Golden &-& -&- &- &- & 8.32 & 0.00  &- & -& 147.07 & 14.45 \\
   
    \bottomrule
    \end{tabular}}
    \caption{Open-ended text generation results, where we also report the evaluation results of the ground truth~(Golden).}
    \label{tab:open_ended_results}
\end{table*}

\subsection{Results}
For all experiments, we compare the performance under the settings of full 2,000 reverse steps and uniformly downsampled 20 steps, aka Respace.
We apply the Minimal Bayesian Risk method~\cite{li2022diffusion} and set candidate size $|\mathcal{S}|$ as 10.

\paragraph{Open-ended Text Generation}
We report the open-ended generation results in Table~\ref{tab:open_ended_results} and observe that our method with 2,000 reverse steps can exceed the DIFFSEQ on most of the evaluation metrics~(except for the Distinct and Lexical Repetition metrics), especially for PPL and Mauve scores that have significant improvement, which means our method can generate high-quality and fluency results.
For downsampled 20 steps, we can find that our method still surpasses the Respace method except for the diversity metrics (D-2 and LR-2) and suffers from a smaller decrease when compared with DIFFUSEQ results of 2,000 steps.
The reason for the high diversity of baselines is that the original DIFFSEQ model or Respace method can generate many meaningless texts, which leads to hard-to-read sentences~(high PPL score).
More details can be referred to Case Study in Appendix~\ref{appendix:case_study}.
Besides, our method also achieves better performance than language models, i.e., CMLM and BART.

\paragraph{Directed Text Generation}
\begin{table}[t]
    \centering
    
    \resizebox{\columnwidth}{!}{
    \begin{tabular}{l|l l l l l l}
    \toprule
    \textbf{Data} & \textbf{Model}  & \textbf{B-2($\uparrow$)} & \textbf{B-4($\uparrow$)} & \textbf{R-2($\uparrow$)} & \textbf{R-L($\uparrow$)}  & \textbf{$\Delta$PPL($\downarrow$)} \\
   
    \midrule
    \multirow{6}{*}{\makecell[l]{WIKI \\ AUTO}} 
    & CMLM &  43.12 & 35.26 & 47.59 & 58.46 &  2.74~(-) \\ 
    & BART &  42.97 & 35.10 & 47.81 & 58.75 &  3.11~(-) \\
    \cmidrule{2-7}      
    &  DIFFSEQ $\dagger$ & 44.02 & 36.08 & 47.18 & 58.43  & \textbf{4.64}~(+) \\
    &  + Ours$\dagger$ & \bf 45.26 & \bf 37.33 & \bf 48.35 & \bf 59.82 & \textbf{2.04}~(-)\\
    \cmidrule{2-7} 
    &  Respace $\ddagger$  & 42.13 & 33.97 & 45.33 & 57.05 &  17.44~(+)\\
    &  + Ours$\ddagger$  & \bf 44.61 & \bf 36.51 & \bf 47.61 & \bf 58.81 & \textbf{3.29}~(+) \\
    
    \midrule
    \multirow{6}{*}{\makecell[l]{QQP}} 
    &  CMLM  & 35.67 & 21.78 & 34.51 & 56.12 & 12.56~(+) \\ 
    & BART  & 33.94 & 20.94 & 33.29 & 54.80 & 8.34~(+) \\
    \cmidrule{2-7}      
    &  DIFFSEQ$\dagger$ & 39.75 & 24.50 & 38.13 & 60.40  & 52.15~(+) \\
    &  + Ours$\dagger$ & \bf 41.74 & \bf 26.27 & \bf 40.56 & \bf 61.88 & \textbf{28.01}~(+)  \\
    \cmidrule{2-7} 
    &  Respace$\ddagger$ & 38.58 & 23.67 & 36.67 & 59.11 & 90.61~(+) \\
    &  + Ours$\ddagger$ & \bf 41.43 & \bf 25.81 & \bf 39.88 & \bf 61.62  & \textbf{35.57}~(+) \\
    \bottomrule
    \end{tabular}}
    \caption{Directed text generation results, where $\dagger$ denotes 2,000 steps, and $\ddagger$ represents 20 steps.}
    \label{tab:directed_generation}
\end{table}
Table~\ref{tab:directed_generation} summarizes the directed text generation results.
We can observe that the improvement in directed text generation tasks is significant that our method achieves better performance than both DIFFSEQ~(2,000 steps) and Respace strategy~(20 steps), especially the PPL score.
We provide more evaluation results of directed generation tasks in Appendix~\ref{appendix:main_results}.

\paragraph{Controllable Text Generation}
\begin{table}[t]
    \centering
    \small
    \resizebox{\columnwidth}{!}{
    \begin{tabular}{l|l c c c}
    \toprule
    \bf Data & \bf Model & \bf Ctrl ($\uparrow$) & \bf  PPL($\downarrow$) & \bf LR-2 ($\downarrow$) \\
    \midrule
    \multirow{5}{*}{\makecell[l]{E2E \\ (Semantic \\ Content)}} & PPLM & 21.03 & 6.04 & 4.18 \\
    \cmidrule{2-5}
    & Diffusion-LM$\dagger$ & 81.46 & 2.52 & \bf 0.08\\
    & + Ours$\dagger$ & \bf 85.06 & \bf 2.38  & 0.68 \\
    \cmidrule{2-5}
    & Respace$\ddagger$ & 75.67 & 2.94  & \bf 0.56\\
    & + Ours$\ddagger$ & \bf 81.87 & \bf 2.66 & 2.18 \\
    \midrule
    \multirow{5}{*}{\makecell[l]{E2E \\ (Syntax \\ Spans)}} & FUDGE & 54.20 & 4.03 & -\\
    \cmidrule{2-5}
    & Diffusion-LM$\dagger$ & 91.12 & 2.52 & \bf 0.35\\
    & + Ours$\dagger$ & \bf 95.33 & \bf 2.33 & 1.54 \\
    \cmidrule{2-5}
    & Respace$\ddagger$ & 82.00 & 2.76 & \bf 0.41\\
    & + Ours$\ddagger$ & \bf 93.15 & \bf 2.68 & 2.39 \\
    \toprule
    \end{tabular}}
    \caption{Results of controllable text generation.}
    \label{tab:controllable_main}
\end{table}
The results of controllable text generation are listed in Table~\ref{tab:controllable_main}, where we follow the official setting to evaluate the PPL\footnote{\url{https://github.com/XiangLi1999/Diffusion-LM/blob/main/train_run.py}}.
Our method can achieve a better control quality than baselines with higher Ctrl scores and generate more fluency results with lower PPL scores but suffers from low diversity.

\subsection{Ablation Study}
We conduct the ablation study on the ROC dataset and set candidate size $|\mathcal{S}|=1$ in this section.

\paragraph{Effect of Distance Penalty}
We first explore the influence of Distance Penalty by adjusting the hyper-parameter $\omega$ and report the results in Table~\ref{tab:penalty_weight}.
We can observe that when the constraint becomes larger, i.e., $\omega$ from 1 to 6, the model can generate more fluent and precise texts but at the cost of diversity.
Besides, we also find that our method can surpass the simple post-training strategy, i.e., $\omega=0$, which means the improvement is brought by the Distance Penalty rather than post-training, which leads to over-fitting on the training data.

\paragraph{Effect of Downsampling Range}
To explore the effect of downsampling range $\mathcal{R}_{k}$ in the training stage, we set the range with the constant and report the results in Table~\ref{tab:ablation_penalty_distance}.
We can observe that as the range becomes larger, i.e., more injected noises, the model can generate more fluent results~(lower PPL) and more precise results~(higher B-2 and R-2) with a smaller sampling range.
Thus, adaptively adjusting the sampling range is essential for making a trade-off among different metrics.

\begin{table}[!tbp]  
    \centering
    \small
    \resizebox{\columnwidth}{!}{
    \begin{tabular}{l|l l l l l l}
    \toprule
    \bf $\gamma$ & \textbf{B-2}($\uparrow$) & \bf R-2 ($\uparrow$)& \textbf{D-2}($\uparrow$) & \bf PPL($\downarrow$) & \bf BS($\uparrow$) & \bf SIM($\downarrow$) \\
    \midrule 
    0 & 8.23 & 3.69 & 25.51 & 99.33 & 52.94 & 16.08 \\ 
    \midrule
    1 & 8.38 & 3.77 & \bf 23.40 & 94.53 & \bf 53.70 & 16.02 \\
    2 & 8.67 & 3.99 & 21.01 & 88.44 & 53.68 & 15.99 \\
    4 & 8.81 & 4.11 & 17.29 & \bf 81.06 & 53.17 & \bf 15.95 \\
    6 & \bf 8.85 & \bf 4.15 & 17.35 & 82.87 & 53.07 & 15.96 \\
    \bottomrule
    \end{tabular}}
    \caption{The influence of penalty weight $\gamma$.}
    \label{tab:penalty_weight}
\end{table}
\begin{table}[t]
    \centering
    \small
    \resizebox{\columnwidth}{!}{
    \begin{tabular}{l | l l l l l l l l}
    \toprule
    \bf Range & \bf Steps & \textbf{B-2}($\uparrow$) & \bf R-2 ($\uparrow$)& \textbf{D-2}($\uparrow$) & \bf PPL($\downarrow$) & \bf BS($\uparrow$) & \bf SIM($\downarrow$) \\
    \midrule
    \multirow{2}{*}{400} & 2,000 & 8.26 & 3.75 & 21.92 & 75.10 & \bf 54.71 & 16.07  \\ 
     & 20 & 8.27 & 3.68 & 23.43 & 89.09 & 54.10 & 16.06  \\
     \midrule
    \multirow{2}{*}{200} & 2,000 & 8.36 & 3.85 & 21.45 & 75.51 & 54.65 & 16.05 \\ 
     & 20 & 8.36 & 3.76 & 23.45 & 93.71 & 53.88 & \bf 16.00 \\
     \midrule 
    \multirow{2}{*}{100} & 2,000 & 8.50 & 3.92 & 21.16 & \bf 73.60 & 54.69 & 16.03 \\ 
     & 20 & 8.38 & 3.77 & 23.40 & 94.53 & 53.70 & 16.02  \\
     \midrule
    \multirow{2}{*}{10} & 2,000 & \bf 8.50 & \bf 3.93 & 20.98 & 74.05 & 54.70 & 16.05  \\ 
     & 20 & 8.42 & 3.84 & \bf 24.24 & 101.62 & 53.60 & 16.06 \\
    \bottomrule
    \end{tabular}}
    \caption{The influence of down-sampling range.}
    \label{tab:ablation_penalty_distance}
\end{table}

\paragraph{Comparison of Sampling Strategies}
\begin{table}[t]
    \small
    \centering
    \resizebox{\columnwidth}{!}{
    \begin{tabular}{l|lccccc}
    \toprule
    \bf Steps & \bf Strategy & \textbf{B-2}($\uparrow$) & \bf R-2 ($\uparrow$)& \textbf{D-2}($\uparrow$) & \bf LR-2 ($\downarrow$) & \bf PPL($\downarrow$) \\
    \midrule
    \makecell[l]{2,000} & - & 8.50 & 3.92 & 21.16 & 0.67 & 73.60 \\
    \midrule
    \multirow{3}{*}{\makecell[l]{200 \\ ($\times 10$)}} & Respace & 8.50 & 3.89 & 20.97 & 0.69 & \bf 73.64 \\
    & DDIM & 8.47 & 3.86 & 17.65 & 1.51 & 77.56 \\
    & ADS  & \bf 8.58 & \bf 3.98 & \bf 21.00 & \bf 0.47 & 73.86 \\
    \midrule
    \multirow{3}{*}{\makecell[l]{20 \\ ($\times 100$)}} & Respace & 8.53 & 3.86 & \bf 20.92 & 0.59 & 79.08 \\
    & DDIM & 7.61 & 3.79 &  15.00 & 3.55 & \bf 73.77 \\
    & ADS & \bf 8.59 & \bf 3.90 & 19.21 & \bf 0.51 & 75.25 \\
    \midrule
    \multirow{3}{*}{\makecell[l]{10 \\ ($\times 200$)}} & Respace &  8.45 & 3.73 & \bf 21.80 & \bf 0.67 & 90.01  \\
    & DDIM & 6.98 & 3.46 & 14.64 & 4.87 & \bf 77.81 \\
    & ADS  & \bf 8.65 & \bf 3.91 & 19.02 & 0.86 & 80.19 \\
    \midrule
    \multirow{3}{*}{\makecell[l]{5 \\ ($\times 400$)}} & Respace &  7.87 & 3.27 & \bf 30.55 & \bf 0.55 & 192.43  \\
    & DDIM & 6.56 & 3.20 & 14.30 & 6.75 & \bf 88.05 \\
    & ADS  & \bf 8.33 & \bf 3.63 & 19.14 & 0.86 & 98.24 \\
    \bottomrule
    \end{tabular}}
    \caption{Comparison of different sampling strategies. The number $\times n$ in brackets illustrates the speedup ratio compared with 2,000 reverse steps.}
    \label{tab:comparison_sampling}
\end{table}

We compare our ADS method with the Respace and the DDIM strategies~(Appendix~\ref{appendix:DDIM}) and can observe that ADS can achieve a better performance~(B-2 and R-2) and generate more fluency texts compared with Respace and more diverse texts compared with DDIM.
Besides, the performance decline of ADS is smaller than the other two strategies, which shows the robustness of ADS~(Appendix~\ref{appendix:robustness_distance_penalty}).

\begin{table}[t]
    \centering
    \small
    \resizebox{\columnwidth}{!}{
    \begin{tabular}{l|l l l l | l l l l }
    \toprule
     \multirow{2}{*}{\bf Metrics} & \multicolumn{4}{c|}{\textbf{2000 steps}} & \multicolumn{4}{c}{ \textbf{20 steps}} \\
     \cmidrule{2-5} \cmidrule{6-9}
     & Win & Loss & Tie & $\zeta$ & Win & Loss & Tie & $\zeta$ \\
    \midrule
    Fluency & 20.6 & 13.8 & 65.6 & 85.9 & 31.1 & 15.0 & 53.9 & 66.6 \\
    Coherence  & 21.7 & 12.8 & 65.5 & 71.0 & 32.8 & 17.2 & 50.0 & 74.0 \\
    Relevance  & 27.8 & 16.1 & 56.1 & 87.8 & 26.7 & 15.6 & 57.7 & 81.7 \\
    \bottomrule
    \end{tabular}}
    \caption{Human evaluation results on two different settings, where $\zeta$ denotes Fleiss' kappa value.}
    \label{tab:human_eval}
\end{table}

\subsection{Human Evaluation}
We compare our method with the vanilla diffusion model on six tasks under 2000 and 20 inference step settings for human evaluation. For each setting, we randomly sample 10 comparison pairs for every task and hire three annotators to give their preferences (win, loss, and tie) for three evaluation criteria: fluency, coherence, and relevance. More details can be referred to Appendix~\ref{appdix:human_evaluation}.
To ensure consistency among the three annotators, we report the Fleiss’ kappa score~\cite{fleiss1971measuring}.
The results are shown in Table~\ref{tab:human_eval}, and we can observe that all the inter-annotator agreements are substantially consistent~($\zeta \in [0.6, 1]$) and our method can achieve better performance than the vanilla diffusion model under both settings. More concretely, as the number of reverse steps decreases, our method can drive the model to generate much better results than the vanilla diffusion model.

\section{Related Work}

\subsection{Text Generation via Diffusion Model}
Denoising diffusion probabilistic models~\cite{ho2020denoising} have shown promising performance on text generation tasks~\cite{yang2022diffusion,li2022diffusion,gong2022diffuseq,austin2021structured}.
There exist two main methodologies, including modeling on \textit{discrete} state spaces or \textit{continuous} state spaces~\cite{sohl2015deep}.
Early works mainly focus on designing one discrete corrupting process on discrete space by introducing the absorbing tokens~\cite{austin2021structured} or transforming the intermediate state into a uniform categorical base distribution~\cite{hoogeboom2021argmax}.
However, such discrete modeling suffers from the scaling of one-hot vectors~\cite{gong2022diffuseq}, which can be only qualified for uncontrollable text generation.
\citet{li2022diffusion} propose the Diffusion-LM, which models the data in the continuous space with one mapping function connecting the continuous space and discrete text space.
\citet{gong2022diffuseq,han2022ssd} combine the diffusion model with iterative NAR model~\cite{gu2019levenshtein} and semi-AR model~\cite{wang2018semi} to further improve the performance in text generation tasks.
Nevertheless, the approaches above all suffer the inefficiency of inference~(reverse process), and the quality of generated text decreases remarkably when applying less denoising steps~\cite{he2022diffusionbert}.

\subsection{Inference Acceleration of Diffusion Model}
One critical drawback of Diffusion Models is that they require many iterations to produce high-quality results.
\citet{song2020denoising} propose one denoising diffusion implicit model (DDIM) and redefine the sampling function to accelerate the generation process.
\citet{jolicoeur2021gotta} devise a faster SDE solver for reverse diffusion processes, and 
\citet{salimans2021progressive} distill a trained deterministic diffusion sampler into a new diffusion model, which only takes half of the sampling steps to generate a full image.
Recent work~\cite{kim2022denoising} also proposes an orthogonal approach Denoising MCMC to accelerate the score-based sampling process of the diffusion model.
Nevertheless, all the methods above are designed for the computer vision field, and the inference acceleration of diffusion for text generation is still unexplored.

\subsection{Exposure Bias of Autoregressive Model}
Exposure Bias is widely recognized as a central challenge in autoregressive models, primarily due to the discrepancies between training and test-time generation, which can result in incremental distortion during the testing phase~\cite{bengio2015scheduled,schmidt2019generalization,he2021exposure}.
To mitigate such issue, three mainstream methods have been adopted, including designing new training objectives~\cite{ranzato2015sequence,shen2016minimum,wiseman2016sequence,zhang2019bridging}, adding regularization terms to standard training objective function~\cite{zhang2019regularizing}, as well as adopting reinforcement learning approaches~\cite{bahdanau2016actor,brakel2017actor} to minimize the expected loss with Minimum Risk Training.

\section{Conclusion}
This work focuses on bridging the training and inference gaps of the diffusion model.
The result of the preliminary study shows that injecting predicted noise into the model can help mitigate the gaps, and the uniform downsampling strategy for inference acceleration harms the model performance.
Thus, we propose two simple yet effective strategies: Distance Penalty and Adaptive Decay Sampling, to mitigate the aforementioned gaps.
Experiments on 6 text generation tasks of 3 different settings show that the model with our methods can achieve better performance and great inference speedup.

\section{Limitation}
Although our method can improve the performance as well as accelerate the inference speed, it suffers from two problems: (1) the diversity of generated results is low compared with language models (LMs) due to the $\textit{clamp}$ sampling strategy, and (2) the diffusion steps of post-tuning stage should stay consistent with the steps in the training stage, and there still exists the gaps between training and inference, i.e., $|\mathcal{T}|=|\mathcal{K}|\neq|\mathcal{T}^{\prime}|$,
To mitigate the aforementioned two issues, we can explore a better post-training or training strategy to mitigate the training-inference gaps further.
In addition, we found that the diffusion model does not perform well in open-ended generation tasks, such as generating incoherent sentences. This is closely related to the drawbacks of NAR models, which have a strong conditional independence assumption. We will attempt to address this issue in the future.

\section*{Ethics Statement}
It is worth noting that all the data used in this paper are publicly available, and we utilize the same evaluation scripts to make sure that all the comparisons are fair.
We have replaced the people names in the corpora with special placeholders to mitigate the problematic biases~\cite{radford2019language} issue of generation results.
Although we have taken some methods to mitigate the problematic biases, such a problem cannot be solved completely.
We urge the users to cautiously apply our methods in the real world and carefully check the generation results.

\section*{Acknowledgement}

This work is supported by the National Science Foundation of China (NSFC No. 62206194 and No. 62106165), the Natural Science Foundation of Jiangsu Province, China (Grant No. BK20220488), and the Project Funded by the Priority Academic Program Development of Jiangsu Higher Education Institutions.
This work is also supported by Beijing Academy of Artificial Intelligence (BAAI).

\bibliography{anthology, custom}
\bibliographystyle{acl_natbib}

\clearpage
\appendix

\section{Preliminary of Diffusion Model}
\label{appendix:detivation_train_object}
In this section, we provide more details of training and inference.
\paragraph{Training Objective}
The training objective of the diffusion model is to maximize the marginal likelihood of distribution $\mathbb{E}_{z_{0}\sim p_{data}}\left[\log p_{\theta}(z_{0})\right]$, and the variational lower bound (VLB) can be written as:
\begin{equation}
\begin{aligned}
& \mathcal{L}_{vlb} = \mathop{\mathbb{E}}\limits_{q(z_{1:T}\mid z_{0})}[\log \frac{q(z_{T}\mid z_{0})}{p_{\theta}(z_{T})} \\
& + \sum\limits_{t=2}^{T}\log \frac{q(z_{t-1}\mid z_{0}, z_{t})}{p_{\theta}(z_{t-1}\mid z_{t})} - \log p_{\theta}(z_{0}\mid z_{1})]
\end{aligned}
\end{equation}
\paragraph{Training}
During the training stage, each intermediate noise $z_{t-1}~(1\leq t \leq T + 1)$ of the forward process can be obtained directly by accumulative multiplication with Equation~\ref{equ:q}:
\begin{equation}
    \label{equ:q_extend}
    q(z_t\mid z_0) = \mathcal{N}(z_t;\sqrt{\bar{\alpha}_t}z_0, (1 - \bar{\alpha}_{t})\mathrm{I}),
\end{equation}
where $\alpha_{t} = 1 - \beta_{t}$ and $\bar{\alpha}_{t} = \prod_{i=1}^{t}\alpha_{i}$.

It is worth noting that, according to the reparameterization method, the value of $1 - \bar{\alpha}_{t}$ denotes the variance of accumulated noise of the current step $z_{t-1}$, i.e., controlling how much noise should be added at the current step.

Combined with Equation~\ref{equ:mu_function}, Equation~\ref{equ:simple_objective} and Equation~\ref{equ:q_extend}, the training process can be referred to Algorithm~\ref{algo:train}~\cite{ho2020denoising}.
\begin{algorithm}[h]
    \caption{Training Process}
    \label{algo:train}
    \begin{algorithmic}[1]
    \Repeat
    \State sample $z_{0}\sim q(z_{0})$
    \State sample $t\sim \mathrm{Uniform}(\{1,\cdots,T\})$
    \State sample $\epsilon\sim \mathcal{N}(0,\mathrm{I})$
    \State  calculate $z_{t} = \sqrt{\bar{\alpha}_{t}}z_0 + \sqrt{1 -\bar{\alpha}_{t}}\epsilon$
    \State gradient descent on $\nabla_{\theta}||\epsilon-\epsilon_{\theta}(z_t, t)||^{2}$
    \Until{converged}
    \end{algorithmic}
\end{algorithm}

\paragraph{Inference}
For the inference stage, there only exists the reverse process, and each intermediate state $z_{t-1}$ is strictly conditioned on the previous history.
It can be summarized into Algorithm~\ref{algo:inference}:
\begin{algorithm}[h]
    \caption{Inference Process}
    \label{algo:inference}
    \begin{algorithmic}[1]
    \State sample $z_{T}\sim \mathcal{N}(0,\mathrm{I})$
    \For{$t\gets T, \cdots, 1$}
        \State sample $\epsilon\sim \mathcal{N}(0,\mathrm{I})$ if $t>1$,else $\epsilon=0$
        \State $z_{t-1} = \frac{1}{\sqrt{\alpha_{t}}}(z_{t} - \frac{\beta_{t}}{\sqrt{1-\bar{\alpha}_{t}}}\epsilon_{\theta}(z_{t}, t)) + \theta_{t}\epsilon$
    \EndFor
     \State \textbf{return} $z_{0}$
    \end{algorithmic}
\end{algorithm}

\section{Diffusion Models for Text Generation}
\label{appendix:embedding}
In this section, we provide more details about Embedding Setting, Clamping Strategy, and Partially Noising Strategy.

\paragraph{Embedding Setting}
As mentioned in Section~\ref{sec:diffsion_for_text_generation}, given the embedding function $\mathcal{E}(\cdot)$, we can map the discrete text into the continuous space or transform the noise back to the discrete text.
Specifically, such mapping strategy, also called \textbf{rounding}~\cite{li2022diffusion}, is achieved by selecting the most probable word in the embedding space by argmax operation: $p_{\theta}(\mathrm{w}\mid z_0) = \prod_{i=1}^{L}p_{\theta}(w_{i}\mid z_{0}^{i})$, where $p_{\theta}(w_{i}\mid z_{0}^{i})$ is a softmax distribution and $z_{0}^{i}$ denotes the $i$-th position of $z_{0}$ distribution.
To train the embedding $\mathcal{E}$, the simplified training objective (Equation~\ref{equ:simple_objective}) should be rewritten as:
\begin{equation}
\label{equ:simple_objective_with_embedding}
\begin{aligned}
    \mathcal{L}^{\prime}_{simple} = & \mathcal{L}_{simple} + \sum\limits_{t=1}^{T}\mathop{\mathbb{E}}\limits_{q(z_{0:T}\mid \mathrm{w})}[||\mathcal{E}(\mathrm{w}) - \\
    & \mu_{\theta}(z_{1}, 1)||^{2} + \log p_{\theta}(\mathrm{w}\mid z_{0})]
\end{aligned}
\end{equation}

\paragraph{Clamping Strategy}
To make the rounding operation more precise, the diffusion model applies the Clamping strategy~\cite{li2022diffusion}, which forces each predicted vector to commit to its nearest word vector through the embedding $\mathcal{E}$ in each reverse step during the inference.
Thus, combined with Equation~\ref{equ:mu_function}, the sampling function of Equation~\ref{equ:sampling} should be rewritten as:
\begin{equation}
    z_{t-1} = \sqrt{\overline{\alpha}}\mathrm{Clamp}(f_{\theta}(z_{t}, t)) + \sqrt{1 - \overline{\alpha}}\epsilon
\end{equation}
Besides, it also approximate the training objective of Equation~\ref{equ:simple_objective_with_embedding} into Equation~\ref{equ:train_objective_for_nlp} by scaling the constants:
\begin{equation}
\label{equ:train_objective_for_nlp}
\begin{aligned}
 & \mathcal{L}_{simple}^{text} = \mathop{\mathbb{E}}\limits_{q(z_{0:T}\mid \mathrm{w})}\Big[
 || \hat{\mu}({z_{T}; z_{0}} ||^2 + \sum\limits^{T}_{t = 2}||\hat{\mu} (z_{t}; z_{0}) \\ 
 & - \mu_\theta( z_t, t)||^2 \Big] + 
 \mathop{\mathbb{E}}\limits_{q(z_{0:1}\mid \mathrm{w})} \big[ ||  
 \mathcal{E}(\mathrm{w})- f_\theta(z_1, 1)||^2 \\
 & - \log p_{\theta}(\mathrm{w} \mid z_0)\big],
\end{aligned}
\end{equation}
where each reverse diffusion step estimates the $z_{0}$ directly rather than $\hat{\mu}({z_{t}, z_{0}})$.

\paragraph{Partially Noising Strategy}
For sequence-to-sequence text generation tasks, \citet{gong2022diffuseq} propose the Partially Noising Strategy that simply perturbs the target text $\mathbf{w}^{x}$ and recovers it conditioned on the source text $\mathbf{w}^{x}$.
More concretely, we concatenate $\mathbf{w}^{x}$ and $\mathbf{w}^{y}$, denoted as $\mathbf{w} ^ {x\bigoplus y}$ and utilize an anchor vector,i.e., $\mathcal{E}(\mathbf{w}^{x})$, to replace the $\mathbf{w}^{x}$ part after each perturbance during the forward diffusion process.
Then, the training objective of the diffusion model can be rewritten as:
\begin{equation}
\label{equ:train_objective_diffseq}
\begin{aligned}
    \mathcal{L}_{seq} = & \mathcal{L}_{simple} + \sum\limits_{t=1}^{T}\mathop{\mathbb{E}}\limits_{q(z_{0:T}\mid \mathrm{w})}[||\mathcal{E}(\mathrm{w} ^ {x\bigoplus y}) - \\
    & \mu_{\theta}(z_{1}, 1)||^{2} + \log p_{\theta}(\mathrm{w} ^ {x\bigoplus y}\mid z_{0})]
\end{aligned}
\end{equation}

\section{Implementation Details}
\label{appendix:implementation_details}
This section provides more details on dataset processing, baseline settings, and evaluation metrics.

\subsection{Dataset Processing}
\label{appendix:implementation_details_datasets}
We provide the statistics of each corpus in Table~\ref{tab:statistic_dataset}.
For the ROC dataset, we mask all the names with special placeholders~\cite{guan2021long} and only keep 4 sentences in the target.
For directed and open-ended generation tasks, we apply the pre-trained tokenizer\footnote{\url{https://huggingface.co/bert-base-uncased}}.
For the E2E dataset, we apply the NLTK package\footnote{\url{https://www.nltk.org/}} for tokenization.

\begin{table}[ht]
    \centering
    \small
    \begin{tabular}{l | l l l}
    \toprule
      \bf Data & \bf \#Train &  \bf \#Valid & \bf \#Test \\
    \midrule
      WIKI-AUTO & 677,751 & 2,038 & 4,972 \\  
      QQP & 114,715 & 878 & 1,091   \\  
      ROC Story  & 88,344 & 4,908 & 4,909   \\ 
      Quasar-T & 116,953 & 2,048 & 10,000 \\ 
      E2E(Semantic) & 333,123 & - & 3,365 \\
      E2E(Syntax) & 41,640 & - & 421 \\
    \bottomrule
    \end{tabular}
    \caption{Statistics of datasets used in our experiments.}
    \label{tab:statistic_dataset}
\end{table}

\subsection{Baselines}
\label{appendix:implementation_details_baselines}
We utilize Diffusion-LM~\cite{li2022diffusion} and DIFFSEQ~\cite{gong2022diffuseq} as the diffusion model baselines, both of which are implemented with transformer model, which contains 12 layers and 12 attention heads.
For a fair comparison, we utilize the language models CMLM~\cite{ghazvininejad2019mask} and BART~\cite{lewis2020bart} that have the same model architecture, i.e., Transformer-based model, and the number of model parameters.
For the CMLM model, we set the iteration steps in inference as 10.
The maximum sequence length is 64 for controllable generation and 128 for directed and open-ended generation tasks.
All the models are trained from scratch, and we set total step $T=2000$ of the diffusion model and apply a square-root noise schedule $\bar\alpha_t = \sqrt{1 - t / T + s}$, where $s$ is a small constant.
We conduct the experiments on 4 NVIDIA A100~(40GB) GPUs~(directed generation and open-ended generation) and 1 NVIDIA TITAN V GPU~(controllable generation\footnote{Due to the limitation of the platform.}). We select the best checkpoint according to the loss on the validation set~(directed generation and open-ended generation) or test the PPL value at the end of each epoch~(controllable generation).
The total training steps and training time~(second) are listed in Table~\ref{tab:training_details_appendix}.
To stay consistent with the baselines, we use the Minimum Bayesian Risk decoding~(MBR) method~\cite{li2022diffusion} in all the experiments, setting the candidate size $|S| = 10$.

\begin{table}[h]
    \centering
    \small  
    \begin{tabular}{l c l}
    \toprule
     \bf Data & \bf Training Step & \bf Time(s) \\
    \midrule
     WIKI-AUTO & 120,000 & 35,978 \\
     QQP & 200,000 & 59,835 \\
     ROC & 120,000 & 35,991 \\
     Quasar-T & 200,000 & 59,911 \\
    E2E(Semantic)$^{*}$ & 120,000 & 15,074 \\
     E2E(Syntax)$^{*}$ & 120,000 & 15,074 \\
    \bottomrule
    \end{tabular}
    \caption{Statistics of training stage, where the datasets denoted with $*$ share the same checkpoint.}
    \label{tab:training_details_appendix}
\end{table} 

\subsection{Evaluation Metrics}
\label{appendix:implementation_details_evaluation}
For Lexical Repetition (LR-$n$) score, we calculate the repetition times $n$ of $k$-gram texts in the generation results and select the hyper-parameter $n$ and $k$ according to the average generation length.
Specifically, we choose $k=4, n=2$ for open-ended generation task, $k=2, n=2$ for directed generation task, and $k=2, n=1$ for controllable generation task.
Besides, for the Semantic Similarity metric, we utilize the Sentence-BERT model\footnote{\url{https://huggingface.co/sentence-transformers/bert-base-nli-mean-tokens}} to compress the whole sentence into a vector and utilize cosine similarity to calculate the distance between two vectors.
We apply the pre-trained GPT-2 model\footnote{\url{https://huggingface.co/gpt2-large}} to calculate the PPL score for open-ended and directed generation tasks as well as utilize the fine-tuned GPT-2 model with E2E dataset for controllable generation task.

\section{Gaps between Training and Inference}
\label{appendix:study_of_gaps}
We provide more detailed preliminary experimental results in this section.

\begin{figure}[t] 
\centering
\subfigure[BLEU-4 scores.]{ 
    \includegraphics[width=0.45\columnwidth]{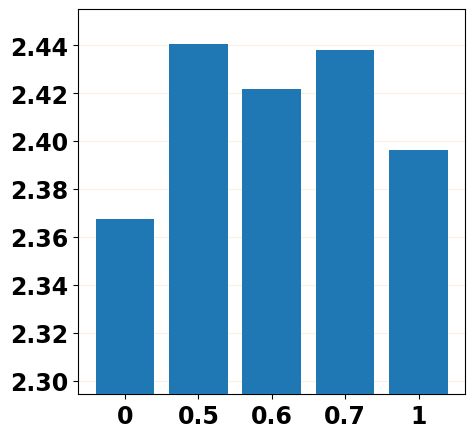}
}
\subfigure[Dist-2 scores.]{ 
    \includegraphics[width=0.45\columnwidth]{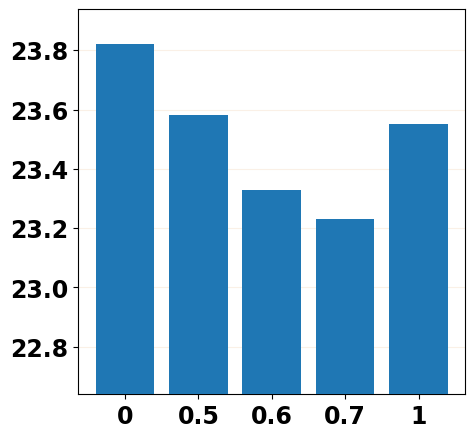}
}
\subfigure[ROUGE-2 scores.] {
    \label{fig:non_uniform_r2}
    \includegraphics[width=0.45\columnwidth]{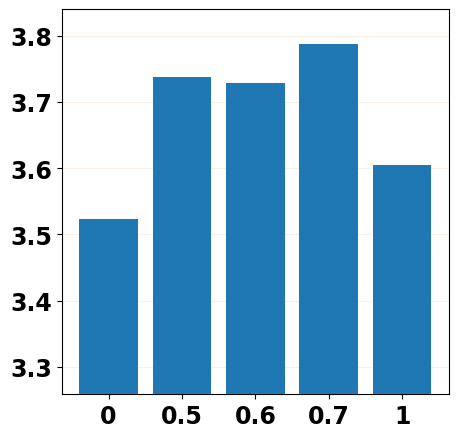}
}
\subfigure[ROUGE-L scores.] { 
    \label{fig:non_uniform_lr2}
    \includegraphics[width=0.45\columnwidth]{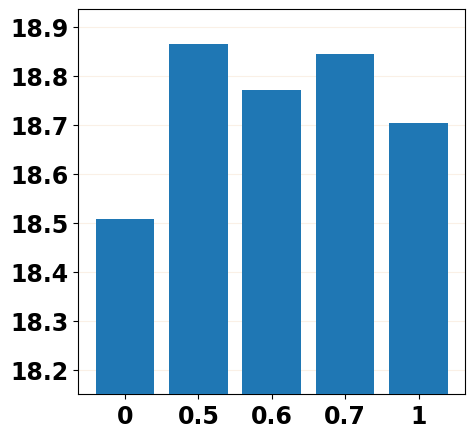}
}

\caption{Evaluation results of noise injection.}
\label{fig:appendix_preliminary_noise}
\end{figure}

\subsection{Training with Predicted Noise}
\label{appendix:noise_injection}
In this section, we provide more details of noise replacement operation.
If we want to replace the conditioned noise $z_{t+1}$ in $p_{\theta}(z_{t}\mid z_{t+1})$ with another noise $z_{t+\delta}$, we can utilize Equation~\ref{equ:p_theta} which transform the probability distribution into the Gaussian distribution, and replace $\mu_{\theta}(z_{t+1}, t+1)$ in Equation~\ref{equ:p_theta} with $\mu_{\theta}(z_{t+\delta}, t+\delta)$.
More experimental results of noise injection are shown in Figure~\ref{fig:appendix_preliminary_noise}.

\subsection{Extensive Trials}
\label{appendix:pseudo_target_selection}
\begin{figure}[t] 
\centering
\subfigure[Trained with 0 step.] { 
    \includegraphics[width=0.45\columnwidth]{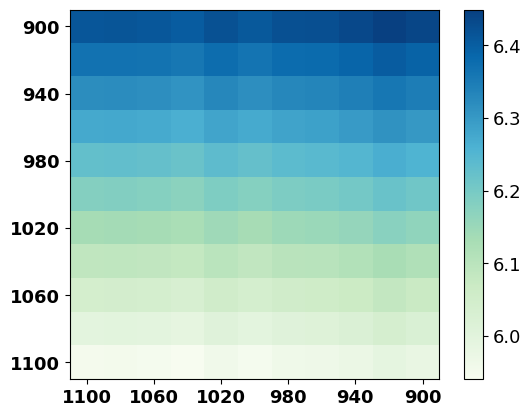}
}
\subfigure[Trained with 4w steps.] { 
    \includegraphics[width=0.45\columnwidth]{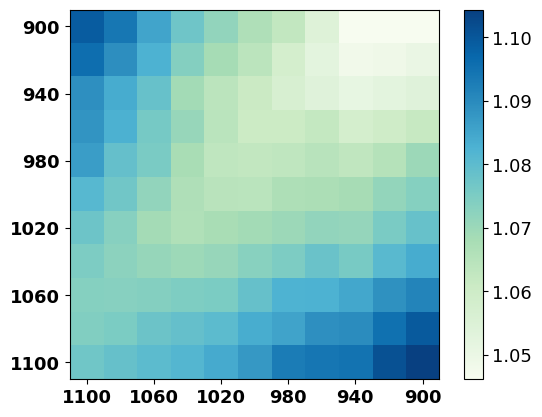}
}
\subfigure[Trained with 8w steps] { 
    \includegraphics[width=0.45\columnwidth]{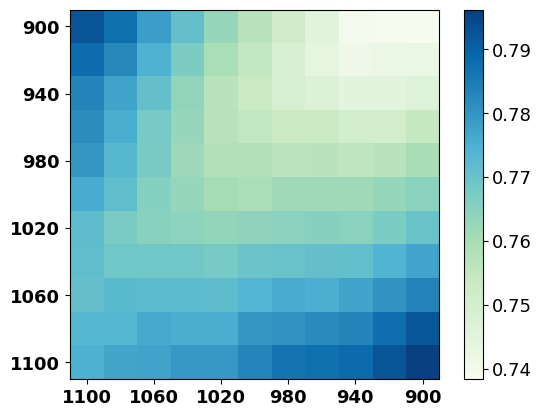}
}
\subfigure[Trained with 12w steps] { 
    \includegraphics[width=0.45\columnwidth]{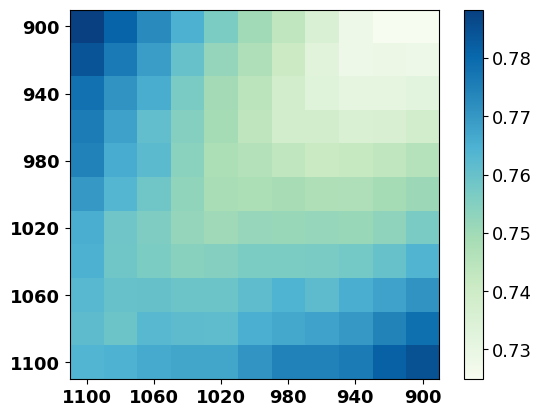}
}
\caption{Euclidean distances between predicted reverse noise and the forward noise, where we set $\delta=20$ and $t=1000$.}
\label{fig:appendix_diff_dis}
\end{figure}

This section shows more results of the distance between the predicted reverse noise with the forward noise.
We plot the result of models trained with 40K, 80K, and 120K steps and show the results of the randomly initialized model, i.e., trained with 0 step, in Figure~\ref{fig:appendix_diff_dis}.
We can observe that the results of the trained models share a similar trend with Figure~\ref{fig:diff_dis}, which indicates that we can inject the predicted reverse noise in the early stage of training~(Appendix~\ref{appendix:distance_penalty}).

\section{Adaptive Decay Sampling}
\label{appendix:adaptive_decay_sampling}
\begin{figure}[t]
    \centering
    \includegraphics[width=1.0\columnwidth]{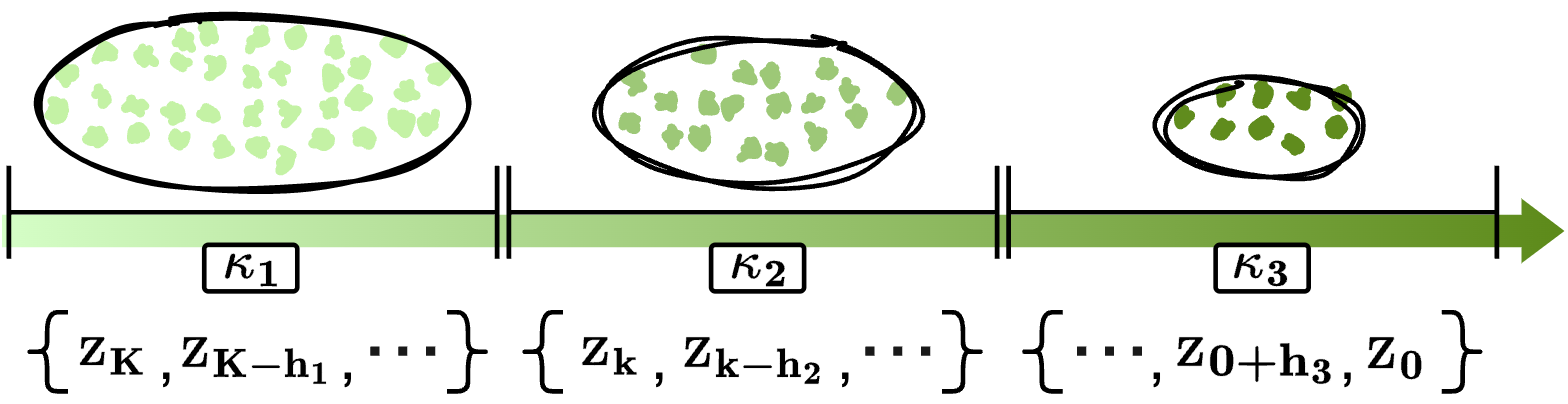}
    \caption{Adaptive Sparse Sampling method.}
    \label{fig:adaptive_sparse_sampling}
\end{figure}

In this section, we introduce the concept of noise scheduler and explain the correlation between the Adaptive Decay Sampling~(ADS) strategy and noise scheduler to help better understand our method.
Besides we also describe the implementation details of the DDIM method.
The overview of the Adaptive Sparse Sampling strategy is shown in Figure~\ref{fig:adaptive_sparse_sampling}, where we split the total down-sampled steps into different subsets for three denoising stages $[\kappa_1, \kappa_2, \kappa_3]$ with weight $\eta_{i}(i\in \{1,2,3\})$

\subsection{Noise Scheduler}
\begin{figure}[ht]
    \centering
    \includegraphics[width=0.9\columnwidth]{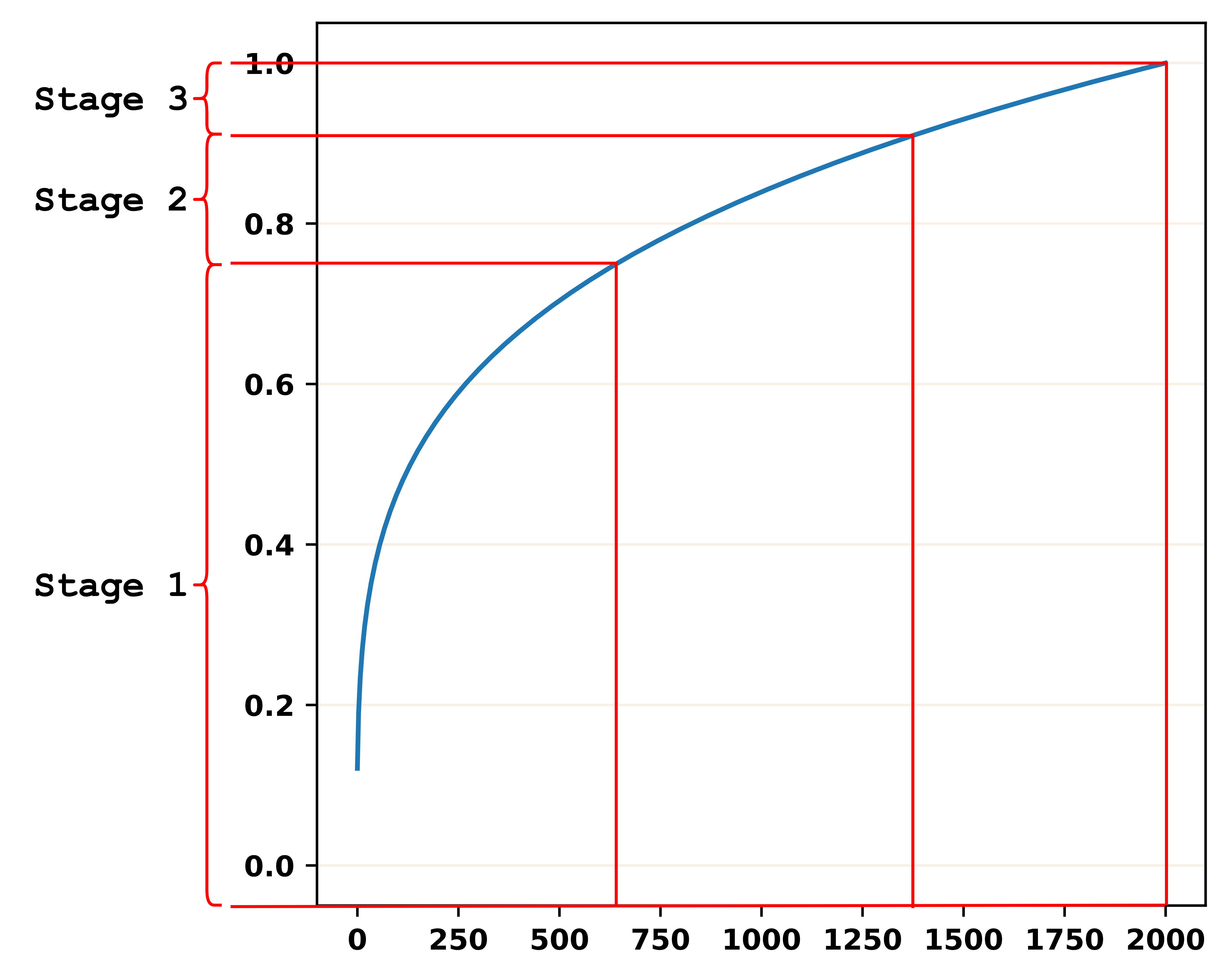}
    \caption{Noise scheduler in the training stage, where the vertical axis represents the weight of added noise and the horizontal axis is the forward diffusion time steps.}
    \label{fig:noise_scheduler_train}
\end{figure}

The noise scheduler controls the amount of noise added to each diffusion forward step parameterized by $\bar{\alpha}_{1:T}$.
As shown in Figure~\ref{fig:noise_scheduler_train}, we plot the sqrt noise scheduler~\cite{li2022diffusion}, which is defined by $\bar{\alpha}_{t} = 1 - \sqrt{t/T + s}$, where $s=1e-4$ is a small constant that simulates the start noise.
We can observe that the noise increases rapidly for the first 500 forward steps and slows down in the latter steps.
When we split the total forward diffusion steps into three stages, we can find the model is trained to solve the high-noise with more steps during the training stage.

\begin{figure}[ht]
    \centering
    \includegraphics[width=0.9\columnwidth]{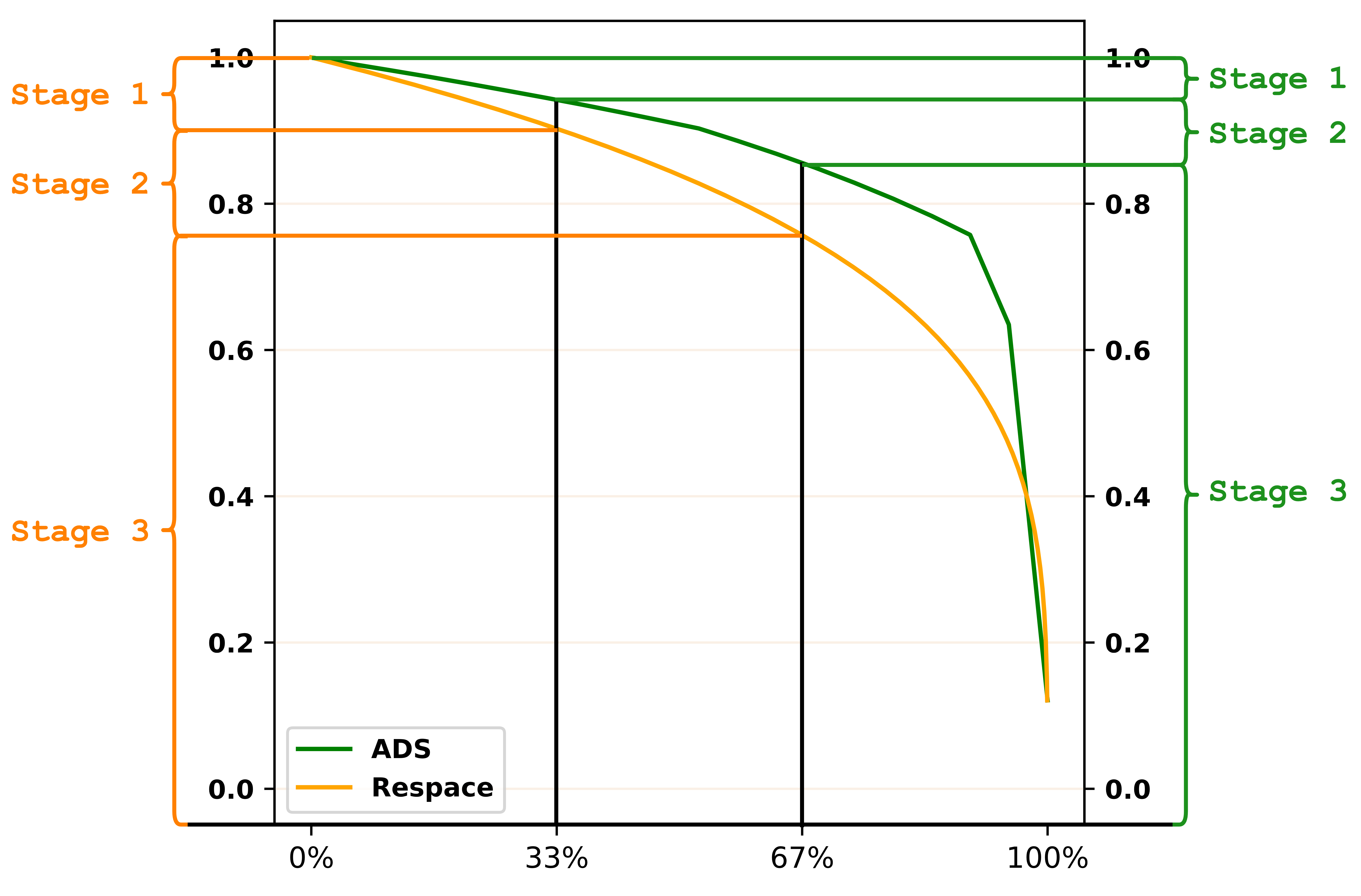}
    \caption{Quantification of remaining noise in inference stage}
    \label{fig:noise_scheduler_inference}
\end{figure}

\subsection{Correlation between ADS and Noise Scheduler}
We also quantify the amount of remaining noise, i.e., the distance between $z_{0}$ and $z_{t}$, in each predicted reverse step $t$ with $\sqrt{1-\bar{\alpha}_{t}}$ and plot the denoising curves in Figure~\ref{fig:noise_scheduler_inference}.
We can observe that the ADS method pays more attention to solving the high-noise compared with Respace strategy, which treats the noise of each stage equally~(\textcolor[RGB]{255,178,100}{yellow curve} v.s. \textcolor[RGB]{29,144,128}{green curve}), and amount of remaining noise decreases rapidly in the third stage~(\textcolor[RGB]{29,144,128}{Stage 3}), which is correspond to $\kappa_3$ in three denoising stages $[\kappa_1, \kappa_2, \kappa_3]$ mentioned in Section~\ref{sec:ADS}.
Besides, Figure~\ref{fig:noise_scheduler_inference} also confirms our preliminary study of sampling strategy in Section~\ref{sec:preliminary_analysis}, i.e., more down-sampled steps for the early denoising stage can improve the model performance.

\subsection{Implementation of DDIM sampling}
\label{appendix:DDIM}
We apply the DDIM sampling strategy~\cite{nichol2021improved} for comparison, which transfers the Markov inference process into the non-Markov one to speed up the inference, i.e., skip some reverse steps during the inference stage.
Given $z_{t-1}$, the sampling function can be written as:
\begin{equation}
    \label{equ:DDIM}
    \begin{aligned}
    z_{t-1} & = \sqrt{\bar{\alpha}_{t-1}}\left(\frac{z_{t}-\sqrt{1-\bar{\alpha}_{t}}f_{\theta}(z_{t}, t)}{\sqrt{\bar{\alpha}_{t}}}\right) \\ & + \sqrt{1-\bar{\alpha}_{t-1} - \sigma^{2}_{t}}f_{\theta}(z_{t}, t) + \sigma_{t}\epsilon_{t},
    \end{aligned}
\end{equation}
where $\epsilon_{t}\sim \mathcal{N}(0, 1)$ and $\sigma_{t}$ is the hyper-parameter.

In this paper, we set $\sigma_{t}=0$ for all time step $t$.

\begin{figure}[t] 
\centering
\subfigure[Change of B-2.] { 
    \includegraphics[width=0.45\columnwidth]{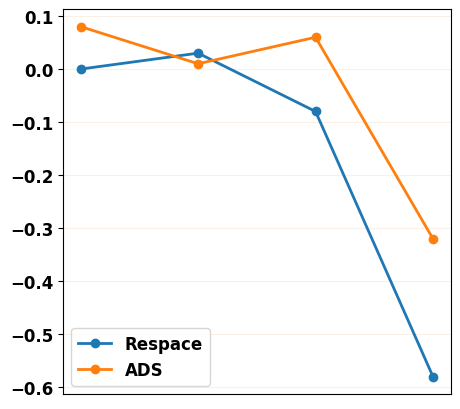}
}
\subfigure[Change of D-2.] { 
    \includegraphics[width=0.45\columnwidth]{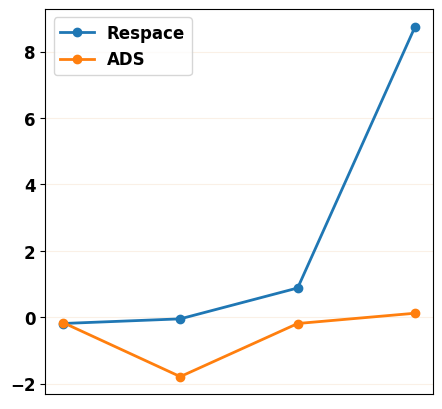}
}
\subfigure[Change of R-2.] { 
    \includegraphics[width=0.45\columnwidth]{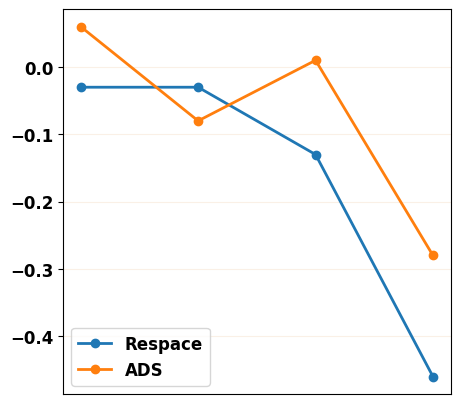}
}
\subfigure[Change of PPL.] { 
    \includegraphics[width=0.45\columnwidth]{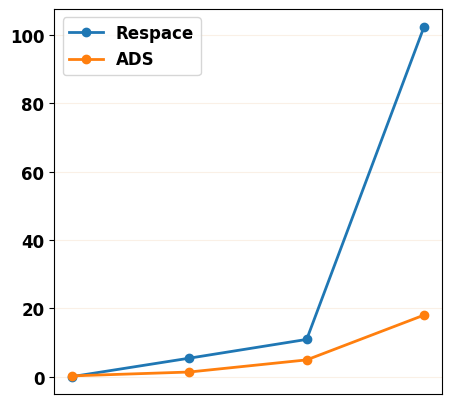}
}
\caption{The change of each evaluation metric between ADS and Respace strategy.}
\label{fig:appendix_robustness}
\end{figure}

\section{Main Result \& Ablation Study}
In this section, we provide more experimental results and implementation details.

\subsection{Evaluation with Fine-tuned GPT-2 Model}
\label{appendix:eval_fine_tuned_gpt2}
We report the Mauve and PPL scores calculated with fine-tuned GPT-2 model for each task in Table~\ref{tab:finetune_ppl_mav}.
Specifically, the GPT-2 model is fine-tuned with the language modeling task on each downstream dataset in 3 epochs and then employed to evaluate the generation results.

\begin{table}[t]
    \centering
    \resizebox{\columnwidth}{!}{
    \begin{tabular}{l|l | cc | cc}
    \toprule
    \textbf{Data} & \textbf{Model} & 
    \textbf{Mav($\uparrow$)$\diamondsuit$} &
    \textbf{Mav($\uparrow$)$\heartsuit$} &
    \textbf{$\Delta$PPL($\downarrow$)$\diamondsuit$} & \textbf{$\Delta$PPL($\downarrow$)$\heartsuit$} \\
   
    \midrule
    \multirow{6}{*}{\makecell[l]{WIKI \\ AUTO}} 
    & CMLM & 99.11 & 98.26 & 2.74~(-) & 5.51 \\ 
    & BART & 99.11 & 98.38 &  3.11~(-) & 7.00 \\
    \cmidrule{2-6}      
    &  DIFFSEQ $\dagger$ & 99.06 & 98.26 & 4.64 & 10.20 \\
    &  + Ours$\dagger$ & 98.76 & 96.76 & 2.04 & 2.40 \\
    \cmidrule{2-6} 
    &  Respace $\ddagger$ & 98.96 & 97.35 & 17.44 & 28.12 \\
    &  + Ours$\ddagger$ & 98.98 & 97.49 & 3.29 & 10.57 \\
    \cmidrule{2-6}
    & Golden & - & - & 110.97 & 77.71 \\
    
    \midrule
    \multirow{6}{*}{\makecell[l]{QQP}} 
    &  CMLM  & 99.58 & 97.95 & 12.56 & 42.24\\ 
    & BART  & 99.67 & 98.39 & 8.34 & 29.51 \\
    \cmidrule{2-6}      
    &  DIFFSEQ$\dagger$ & 98.40 & 94.47 & 52.15 & 55.73\\
    &  + Ours$\dagger$ & 98.84 & 96.03 & 28.01 & 30.17 \\
    \cmidrule{2-6} 
    &  Respace$\ddagger$ & 97.70 & 90.63 & 90.61 & 88.60 \\
    &  + Ours$\ddagger$ & 98.54 & 95.63 & 35.57 & 38.96 \\
    \cmidrule{2-6}
    & Golden & - & - & 40.11 & 41.02\\

    \midrule
    \multirow{7}{*}{ROC} 
    & CMLM &  2.73 &  8.72 & 13.45 & 85.15 \\ 
    & BART & 70.64 & 74.49 & 2.95 & 9.14 \\
    \cmidrule{2-6}    
    &  DIFFSEQ$\dagger$ & 34.45 & 62.21 & 49.44 & 129.03 \\
    &  + Ours$\dagger$ & 41.56 & 64.00 & 33.56 & 101.86 \\
    \cmidrule{2-6} 
    &  Respace$\ddagger$ & 25.37 & 56.06 & 56.32 & 139.87 \\
    &   + Ours$\ddagger$ & 28.06 & 54.26 & 48.76 & 127.72 \\
    \cmidrule{2-6}  
    &  Golden & - & - & 29.72 & 21.69 \\

    \midrule
    \multirow{7}{*}{Quasar-T} 
    & CMLM &  1.96 &  0.96 & 88.37 & 159.33 \\ 
    & BART & 3.09 & 3.07 & 0.75 & 68.63 \\
    \cmidrule{2-6}    
    &  DIFFSEQ$\dagger$ & 4.68 & 4.36 & 95.68 & 44.46 \\
    &  + Ours$\dagger$ & 10.91 & 5.21 & 58.68 & 21.24 \\
    \cmidrule{2-6} 
    &  Respace$\ddagger$ & 4.53 & 4.72 & 169.75 & 79.14 \\
    &  + Ours$\ddagger$ & 5.41 & 5.41 & 96.83 & 43.26 \\
    \cmidrule{2-6}  
    &  Golden & -&- &147.07 & 71.99 \\

    \bottomrule
    \end{tabular}}
    \caption{Comparison of Mauve and PPL scores between fine-tuned model and pretrained model, where $\dagger$ denotes 2,000 steps, and $\ddagger$ represents 20 steps. $\diamondsuit$ and $\heartsuit$ represents the fine-tuned model and pre-trained model respectively. }
    \label{tab:finetune_ppl_mav}
\end{table}

\begin{table}[t]
    \centering
    \small
    \resizebox{\columnwidth}{!}{
    \begin{tabular}{l | l l | l l}
    \toprule
     \multirow{2}{*}{Dataset} & \multicolumn{2}{c|}{2000 steps} & \multicolumn{2}{c}{20 steps} \\
     \cmidrule{2-3} \cmidrule{4-5}
     & \#Num & \#Len & \#Num & \#Len \\
     \midrule
     ROC  & 10 & 40.4 & 10 & 41.1 \\
     Qusar-T & 10 & 12.7  & 10 & 13.3 \\
     WIKI AUTO & 10 & 27.5 & 10 & 26.0\\
     QQP & 10 & 11.2 & 10 & 10.3\\
     E2E(Semantic) & 10 & 22.0 & 10 & 26.5\\
     E2E(Syntax) & 10 & 26.7 & 10 & 27.2\\
     \bottomrule
    \end{tabular}}
    \caption{Statistic of human evaluation data, where \#Num denotes the number of cases and \#Len denotes the average length of sampled instances for each task.}
    \label{tab:human_eval_cases}
\end{table}

\subsection{Directed Generation Results}
\label{appendix:main_results}
We provide the full evaluation results of directed generation tasks in Table~\ref{tab:appendix_direct_generation}.
\begin{table*}[t]
    \centering
    \small
    \resizebox{\textwidth}{!}{
    \begin{tabular}{ll | l l l l l l l l  l}
    \toprule
    \textbf{Data} & \textbf{Model}  & \textbf{Step} & \textbf{B-2($\uparrow$)} & \textbf{B-4($\uparrow$)} & \textbf{R-2($\uparrow$)} & \textbf{R-L($\uparrow$)} & \textbf{LR-2($\downarrow$)} & \textbf{BS($\uparrow$)} & \textbf{Mav($\uparrow$)} & \textbf{$\Delta$ PPL($\downarrow$)} \\
   
    \midrule
    \multirow{7}{*}{\makecell[l]{WIKI-AUTO}} 
    & CMLM & 10 & 43.12 & 35.26 & 47.59 & 58.46 &  2.94 & 81.83 & 98.19 &  2.74~(-)\\ 
    & BART & - & 42.97 & 35.10 & 47.81 & 58.75 &  2.22 & \underline{81.98} & \underline{98.38} &  \underline{3.11~(-)}\\
    \cmidrule{2-11}      
    &  DIFFSEQ   & 2,000 & 44.02 & 36.08 & 47.18 & 58.43  & \bf 1.65 & 81.27 & \bf 98.26 & 4.64~(+) \\
    &  + Ours$\dagger$ & 2,000 & \bf 45.26 & \bf 37.33 & \bf 48.35 & \bf 59.28  & 2.00 & \bf 81.88 & 96.76 & \bf 2.04~(-) \\
    \cmidrule{2-11} 
    &  + Respace & 20  & 42.13 & 33.97 & 45.33 & 57.05   & \bf 1.37 & 79.94 &  97.35 &  17.44~(+) \\
    &  + Ours$\ddagger$ & 20 & \bf 44.61 & \bf 36.51 & \bf 47.61 & \bf 58.81 &  1.65 & \bf 81.42 & \bf 97.49 & \bf 3.29~(+) \\
    \cmidrule{2-11} 
    &  Golden & -&-&- &- &- & 1.95 &- &- & 77.71 \\
    \midrule
    \multirow{7}{*}{\makecell[l]{QQP}} 
    &  CMLM & 10 & 35.67 & 21.78 & 34.51 & 56.12  & 0.04 & 82.86 & 97.75 &  12.56~(+)\\ 
    & BART & - & 33.94 & 20.94 & 33.29 & 54.80  & 0.28 & 82.28 & \underline{98.39} &  \underline{8.34~(+)} \\
    \cmidrule{2-11}      
    &  DIFFSEQ   & 2,000 & 39.75 & 24.50 & 38.13 & 60.40  & 0.09 & 83.41 & 94.47 & 52.15~(+) \\
    &  + Ours$\dagger$ & 2,000 & \bf 41.74 & \bf 26.27 & \bf 40.56 & \bf 61.88 & \bf 0.00 & \bf 84.72 & \bf 96.03 & \bf 28.01~(+) \\
    \cmidrule{2-11} 
    &  + Respace & 20  & 38.58 & 23.67 & 36.67 & 59.11 &  \bf 0.00 & 82.16 & 90.63 & 90.61~(+) \\
    &  + Ours$\ddagger$ & 20 & \bf 41.43 & \bf 25.81 & \bf 39.88 & \bf 61.62  & \bf 0.00 & \bf 84.35 & \bf 95.63 & \bf 35.57~(+) \\
    \cmidrule{2-11} 
    &  Golden &- &- &- &- &- & 0.18 &- &- & 83.84 \\
    \bottomrule
    \end{tabular}}
    \caption{Directed text generation results.}
    \label{tab:appendix_direct_generation}
\end{table*}

\subsection{Sampling Strategy Comparison}
We provide the full results of different sampling strategies on ROC and WIKI-AUTO datasets as well as report the inference speed\footnote{When calculating the inference speed, we set the length of generated results to the same and control the batch size as 1.} in Table~\ref{tab:acclerating_comparison}.
\begin{table*}[t]
    \centering
    \small
    \resizebox{\textwidth}{!}{
    \begin{tabular}{l l | l l l l l | l l l l l | l l}
    \toprule
    \bf \multirow{2}{*}{Steps} & \bf \multirow{2}{*}{Methods} & \multicolumn{5}{c}{\textbf{Story Generation}} & \multicolumn{5}{|c|}{\textbf{Text Simplification}} & \bf \multirow{2}{*}{T/s} &\bf \multirow{2}{*}{I/s} \\
    \cmidrule{3-12} 
    & & \bf B-2 & \bf D-2 & \textbf{PPL} & \bf Sim & \bf BS & \bf B-2 & \bf R-2 & \bf R-L & \textbf{PPL}& \bf BS &  \\
    \midrule
    2000 & origin & 8.09 & 23.82 & 90.78 & 16.12 & 53.98 & 35.48 & 38.35 & 51.39 & 119.84 & 76.42 & 6.51 & 0.05 \\
    \midrule
    \multirow{3}{*}{200}
     &  Respace & 8.08 & \bf 23.95 & 92.23 & 16.13 & 53.86 & 36.13 & 38.89 & 51.94& 119.86 & 76.66 & 63.77 & 0.49 \\
     &  DDIM & 8.22 & 20.86 & 95.65 & 16.03 & 52.87 & 25.52 & 31.67 & 42.54 & 102.09 & 66.43 & 62.33 & 0.48 \\
     &  Ours & \bf 8.58 & 21.00 & \bf 73.86 & \bf 16.02 & \bf 54.63 & \bf 39.70 & \bf 43.17 & \bf 55.15 & \bf 96.07 & \bf 78.88 & 61.67 & 0.48 \\
     \midrule
     \multirow{3}{*}{20} 
     &  Respace & 8.07 & \bf 24.21 & 98.33 & 16.14 & 53.60 & 37.26 & 39.21 & 52.72 & 130.29 & 76.41 & 622.09 & 4.86 \\
     &  DDIM & 7.57 & 19.66 & 98.77 &  16.01 & 51.41 & 10.37 & 16.83 & 25.85 & 116.01 & 50.93 & 582.53 & 4.55 \\
     &  Ours & \bf 8.59 & 19.21 & \bf 75.25 & \bf 15.95 & \bf 54.08 & \bf 41.62 & \bf 44.33 & \bf 56.51 & \bf 101.74 & \bf 79.43 & 604.35 & 4.72 \\
     \midrule
     \multirow{3}{*}{10} 
     &  Respace & 8.35 & \bf 23.91 & 115.21 & 16.14 & 53.06 & 36.75 & 38.32 & 51.75 & \bf 143.78 & 75.27 & 1145.70 & 8.95 \\
     &  DDIM & 6.99 & 19.62 & 107.52 & 16.08 & 50.38 & 8.73 & 12.89 & 21.86 & 152.67 & 48.74 & 1173.29 & 9.16 \\
     &  Ours & \bf 8.65 & 19.02 & \bf 80.19 & \bf 15.88 & \bf 53.78 & \bf 39.57 & \bf 41.11 & \bf 54.72 & 150.97 & \bf 77.42 & 1200.55 & 9.37 \\
     \midrule
     \multirow{3}{*}{5} 
     &  Respace & 7.21 & \bf 35.49 & 252.06 & 16.34 & 50.62 & 34.05 & 34.75 & 48.89 & 207.79 & 72.14 & 2240.38 & 17.50  \\
     &  DDIM & 6.51 & 21.18 & 134.36 & 16.22 & 48.97 & 6.57 & 8.13 & 17.26 & 255.86 & 45.16 & 2257.06 & 17.63 \\
     &  Ours & \bf 8.33 & 19.14 & \bf 98.24 & \bf 15.96 & \bf 52.62 & \bf 38.42 & \bf 39.70 & \bf 53.50 & \bf 171.33 & \bf 76.03 & 2217.23 & 17.32 \\
    \bottomrule
    \end{tabular}}
    \caption{Comparison among different sampling strategies on ROC and WIKI-AUTO datasets, where ``T/s'' denotes Tokens/s and ``I/s'' means Instances/s.}
    \label{tab:acclerating_comparison}
\end{table*}

\subsection{Speed up the Training}
\label{appendix:distance_penalty}
To save the total training time, we explore the insights of post-training the diffusion model with the Distance Penalty method from its early training stage rather than from the end of its training stage.
We conduct the experiment on the ROC dataset and set the candidate size $|S|$ of MBR as 1 for convenience.
As shown in Table~\ref{tab:continual_training}, we can observe that post-tuning with the Distance Penalty method can bring massive improvement to the diffusion model, and it can still achieve a great performance when post-training the model with few warm-up training steps, i.e., 40K(Start) + 30K(Post).
Besides, the improvement will be more significant when post-training the model with more training steps.

\begin{table*}[t]
    \centering
    \small
    \resizebox{\textwidth}{!}{
    \begin{tabular}{l| l l l l l l l l l c l l}
    \toprule
    \textbf{Begin} & \textbf{Post} & \textbf{Steps}& \textbf{B-2$(\uparrow$)} & \textbf{B-4$(\uparrow$)} & \textbf{R-2$(\uparrow$)} & \textbf{R-L$(\uparrow$)} & \textbf{D-2$(\uparrow$)} &  \textbf{LR-2$(\downarrow$)} & \textbf{BS$(\uparrow$)} & \textbf{Mav$(\uparrow$)} & \textbf{PPL$(\downarrow)$} & \textbf{ SIM$(\downarrow$)}\\
    \midrule
    \multirow{4}{*}{40K} & \multirow{2}{*}{0} & 20 & 7.35 & 2.11 & 2.84 & 17.60 & \bf 21.62 & \bf 0.16 & 51.78 & 4.13 & 133.40 & 16.41 \\
     & & 2,000 & 7.37 & 2.14 & 0.30 & 17.69 & 21.10 & 0.35 & 52.27 & 6.69 & 116.67 & 16.30
       \\
     \cmidrule{2-13}
     & \multirow{2}{*}{30K} & 20 & \bf 8.65 & \bf 2.53 & 3.97 & \bf 19.57 & 17.75 & 1.96 & 53.91 & 19.97 & 73.64 & \bf 16.04 \\
     & & 2,000 & 8.49 & 2.49 & \bf 4.03 & 19.49 & 17.50 & 2.02 & \bf 54.29 & \bf 29.71 & \bf 65.92 & 16.05 \\
     \midrule
    \multirow{4}{*}{80K} & \multirow{2}{*}{0} & 20 & 8.56 & 2.50 & 3.92 & 19.24 & \bf 20.77 & \bf 1.12 & 54.38 & 35.54 & 77.77 & 16.00 \\
     & & 2,000 & 8.44 & 2.48 & 4.01 & 19.15 & 20.36 & 1.33 & 54.78 & \bf 46.89 & 68.70 & 15.99 \\
     \cmidrule{2-13}
     & \multirow{2}{*}{30K} & 20 & \bf 8.79 & \bf 2.63 & 4.14 & \bf 19.59 & 19.47 & 1.35 & 54.57 & 34.02 & 70.98 & 15.99 \\
     & & 2,000 & 8.77 & 2.62 & \bf 4.27 & 19.57 & 19.48 & 1.26 & \bf 55.08 & 45.90 & \bf 63.93 & \bf 15.97 \\

     \midrule
    \multirow{4}{*}{120K} & \multirow{2}{*}{0} & 20 & 8.37 & 2.46 & 3.79 & 19.01 & \bf 22.23 & 0.96 & 54.45 & 37.67 & 80.98 & 16.08 \\
     & & 2,000 & 8.37 & 2.48 & 3.93 & 19.01 & 21.86 & \bf 0.80 & 54.85 & \bf 54.79 & 71.98 & 16.07 \\
     \cmidrule{2-13}
     & \multirow{2}{*}{30K} & 20 & \bf 8.68 & \bf 2.58 & 4.03 & \bf 19.44 & 20.54 & 1.28 & 54.55 & 35.23 & 74.10 & 16.03   \\
     & & 2,000 & 8.62 & \bf 2.58 & \bf 4.10 & 19.37 & 20.43 & 1.28 & \bf 54.97 & 45.59 & \bf 66.90 & \bf 16.00 \\

    \midrule
    \multirow{4}{*}{240K} & \multirow{2}{*}{0} & 20 & 8.07 & 2.37 & 3.50 & 18.53 & \bf 24.21 & \bf 0.31 & 53.60 & 26.39 & 98.33 & 16.14 \\
     & & 2,000 & 8.09 & 2.38 & 3.61 & 18.51 & 23.82 & 0.57 & 53.98 & 34.06 & 90.78 & 16.12 \\
     \cmidrule{2-13}
     & \multirow{2}{*}{30K} & 20 & \bf 8.53 & 2.50 & 3.86 & \bf 19.23 & 20.92 & 0.59 & 54.30 & 29.28 & 79.08 & 16.04 \\
     & & 2,000 & 8.50 & \bf 2.51 & \bf 3.92  & 19.19 & 21.16 & 0.67 & \bf 54.69 & \bf 39.94 & \bf 73.60 & \bf 16.03 \\
    \bottomrule
    \end{tabular}}
    \caption{Evaluation results of post-training based on models with different training steps, where ``Begin'' denotes the training steps and ``Post'' denotes the post-training steps.}
    \label{tab:continual_training}
\end{table*}


\subsection{Robustness of Adaptive Decay Sampling}
\label{appendix:robustness_distance_penalty}
To reflect the decrease of each evaluation metric along with fewer inference steps more clearly, we plot the rate of change for each metric in Figure~\ref{fig:appendix_robustness}.
We can find that the change rate of our ADS strategy is lower than the Respace strategy, which means our method has better robustness as the number of down-sampled steps decreases.

\section{Human Evaluation}
\label{appdix:human_evaluation}
\begin{figure*}[t]
    \centering
    \includegraphics[width=1.0\textwidth]{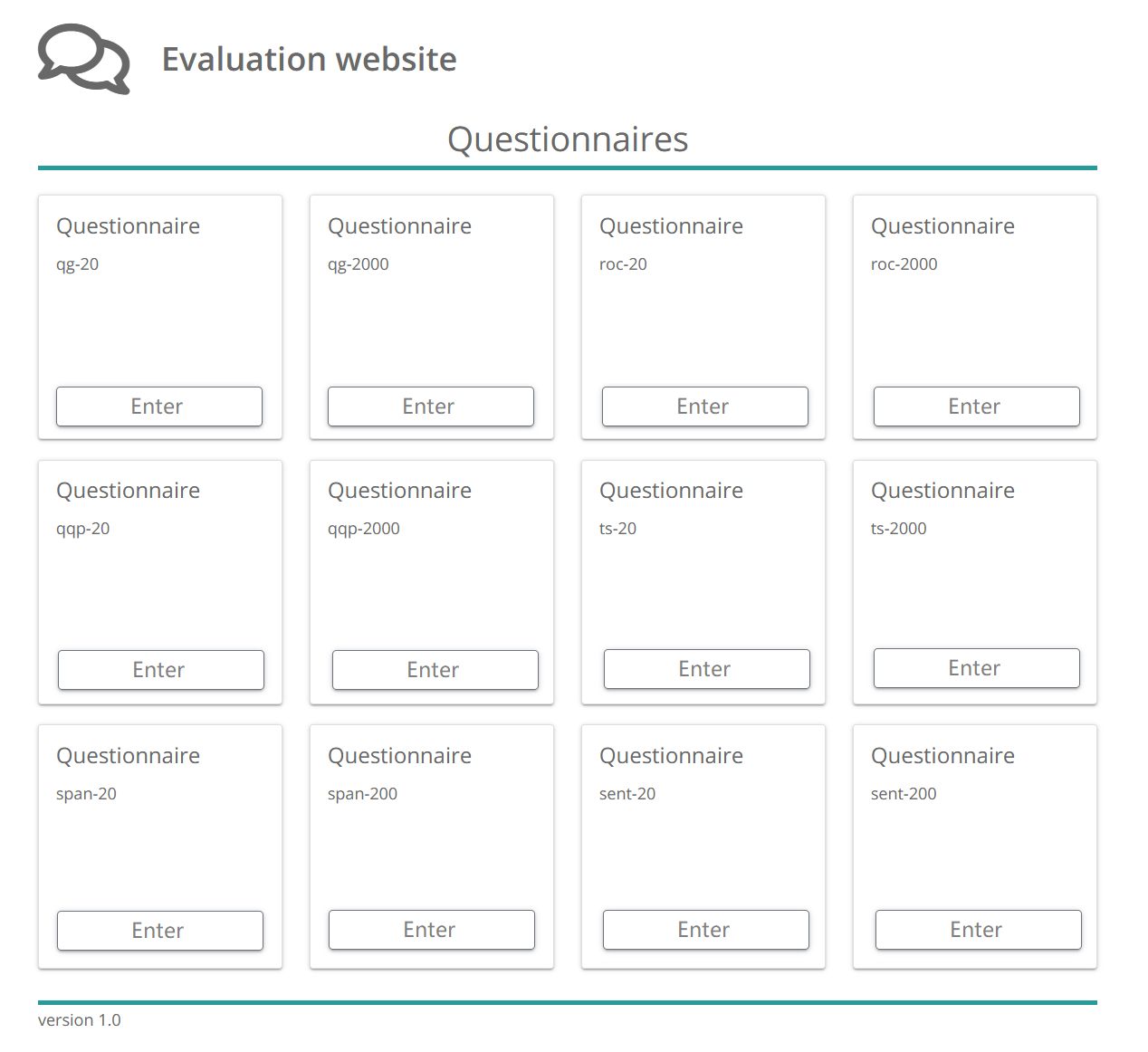}
    \caption{Interface of human evaluation website.}
    \label{fig:huamn_eval_interface}
\end{figure*}

\begin{figure*}[t]
    \centering
    \includegraphics[width=1.0\textwidth]{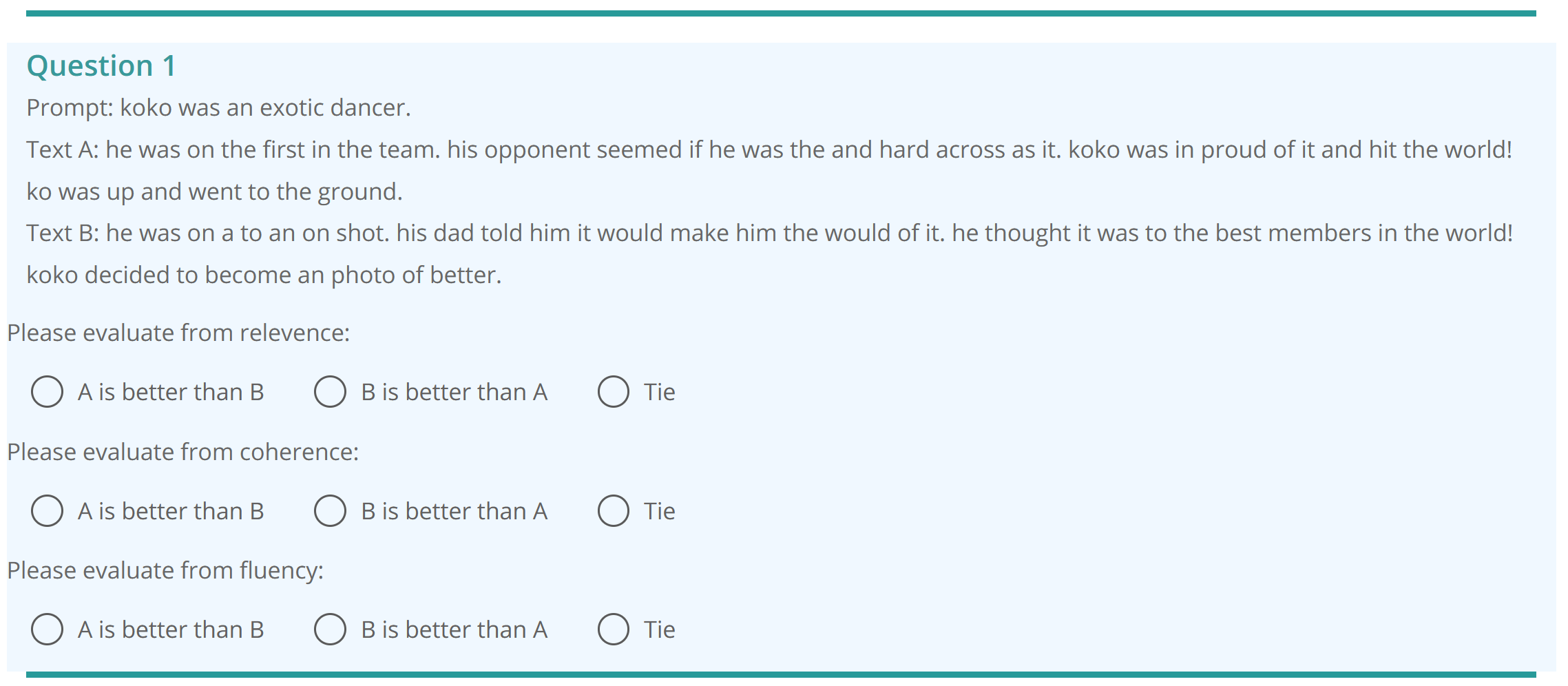}
    \caption{Example of one comparison pair in the human evaluation website.}
    \label{fig:huamn_eval_comparison}
\end{figure*}

We show the statistic of human evaluation data in Table~\ref{tab:human_eval_cases} and human evaluation interface in Figure~\ref{fig:huamn_eval_interface} and~\ref{fig:huamn_eval_comparison}. We build the human evaluation interface with the open-source python web library Django~\footnote{\url{https://www.djangoproject.com}}.
As shown in Figure~\ref{fig:huamn_eval_comparison}, during the evaluation, each comparison pair contains one prompt and two corresponding outputs generated from two different models. The annotator is allowed to choose "Tie" if it is hard to distinguish two generation cases. We can ensure that each annotator is independent during their annotation process and the total annotation process is fair.
We hired three annotators and payed each annotator \$ 0.05 for comparing each pair. The payment is reasonable considering that it would cost average 30 seconds for an annotator to finish a comparison.

\section{Case Study}
\label{appendix:case_study}
In this section, we present part of the generated results of each task for better illustration.
We randomly select the cases generated by the diffusion model of 2,000 steps and 20 steps, and the diffusion model post-trained with the Distance Penalty method~(2,000 steps) and ADS strategy~(20 steps). 
For clear representation, we utilize \textcolor{green}{green color} denotes the key phrase related to the prompt, \textcolor{red}{red color} locates the phrase contradicts the prompt, and \textcolor{blue}{blue color} highlights the serious grammatical errors.
We can observe that, with our methods, the model can generate more fluency and high-quality texts, while the original model generates many repetitions or meaningless tokens.
It is worth noting that the language model~(pre-trained GPT-2) may allocate a good PPL score to the sentence with many repetition tokens, and those texts with massive meaning-less tokens or hard-to-read sentences may achieve a better Distinct and Lexical Repetition score.

\begin{table*}[t]
\centering
\small
\resizebox{\textwidth}{!}{
\begin{tabular}{c | l | l }
\toprule
Case & Type & Text \\
\midrule
\multirow{1}{*}{1} & Prompt & i was on my laptop computer. \\
    \cmidrule{2-3}
     & Origin (2000) & \makecell[l]{all of a sudden my computer stopped working on the table. i was scared because it ripped so i ended up buying it. \\ i went to the store to get a new one. after half of my computer, the screen were fine.} \\
     \cmidrule{2-3}
     & + Ours (2000) & \makecell[l]{all of a sudden my computer stopped working. the screen was broken. \\ it was so frustrated i had to get a replacement. luckily i was able to fix.} \\
     \cmidrule{2-3}
     & + Respace (20) & \makecell[l]{all of a sudden my computer \textcolor{blue}{stopped and stopped} working. i was very shocked and realized it was my computer. \\ i called to the coffee again and had to buy a new one. after paying the work, my new computer broke.}\\
     \cmidrule{2-3}
     & + Ours (20) & \makecell[l]{all of a sudden my computer stopped. it broke the screen. \\ it was not flying and i took it back to the store to fix it. luckily, it was fixed.}\\
     \cmidrule{2-3}
     & Gold & \makecell[l]{i was writing a report for work. i was typing it up very quickly. my hands began to hurt. i took a break from the report.} \\
\midrule
\multirow{1}{*}{2} & Prompt & alex has a big exam on friday. \\
    \cmidrule{2-3}
     & Origin (2000) & \makecell[l]{he has been studying all week. he got out of class. he got into class instead. his teacher gave the test and knew and made \\ an a.} \\
     \cmidrule{2-3}
     & + Ours (2000) & \makecell[l]{he has been studying all week. he got to class. when he got to class, his teacher handed out the test. alex was thrilled.} \\
     \cmidrule{2-3}
     & + Respace (20) &\makecell[l]{he has been studying all week. he got to class ready. he got into class. and his teacher got it. alex was chosen nervous  \\and a slept.} \\
     \cmidrule{2-3}
     & + Ours (20) &\makecell[l]{he has been studying all week. he stayed up all night. he got to class days early. alex took the test and was so happy!} \\
     \cmidrule{2-3}
     & Gold & \makecell[l]{he studied day in and day out. however, he didn't take any breaks.\\ this caused him to second guess his answers. he ended up failing the test because he over study.} \\
\midrule
\multirow{1}{*}{3} & Prompt & i went to the mall yesterday. \\
    \cmidrule{2-3}
     & Origin (2000) & \makecell[l]{i thought there was a sign sale. i ran off at the store. i found out it was one that sale. i was so upset.} \\
     \cmidrule{2-3}
     & + Ours (2000) & \makecell[l]{i thought there was a sale. so i walked up at the store. it was on sale. i was so upset.}\\
     \cmidrule{2-3}
     & + Respace (20) & \makecell[l]{i thought there was a sale. so i got up at the store. there was one sale. i was thrilled.} \\
     \cmidrule{2-3}
     & + Ours (20) & \makecell[l]{i thought there was a sale. so i went to the electronics store. there was a good sale. i was pleased with them.} \\
     \cmidrule{2-3}
     & Gold & \makecell[l]{i tried on some lip gloss. it looked pretty. yet it burned my lips. i had to wipe it off.} \\
\midrule
\multirow{1}{*}{4} & Prompt & howard had his first baseball game. \\
    \cmidrule{2-3}
     & Origin (2000) & \makecell[l]{he was excited. when he went on the pitcher. he made sure he started to \textcolor{blue}{practice he would get}. he did it in a display.} \\
     \cmidrule{2-3}
     & + Ours (2000) & \makecell[l]{he was very excited when he went on the game. he was nervous. at the moment \textcolor{blue}{he did he put it back}. he ended up \\ winning a day.}\\
     \cmidrule{2-3}
     & + Respace (20) & \makecell[l]{he was excited. when he went to the game. he got up at first. he got to beat - game. he did up a time. } \\
     \cmidrule{2-3}
     & + Ours (20) & \makecell[l]{he was very excited to try out in the game. he was nervous. when he started it, he found out he did. he did it in the park.} \\
     \cmidrule{2-3}
     & Gold & \makecell[l]{he was very nervous. he ends up striking out. he left the game disheartened. his parents gave him cake to cheer him up.} \\

\midrule
\multirow{1}{*}{5} & Prompt & sara wanted to try out for the dance team. \\
    \cmidrule{2-3}
     & Origin (2000) & \makecell[l]{she was both nervous and excited. she often tried it but she didn't know how try it. she finally decided to try out. she \\ loved it, \textcolor{blue}{but and she both loved it}.} \\
     \cmidrule{2-3}
     & + Ours (2000) & \makecell[l]{she was both nervous and excited. she had a years left and didn't want to do it. she finally decided to try out. she made it \\ and was glad she made it.}\\
     \cmidrule{2-3}
     & + Respace (20) & \makecell[l]{ she was both nervous and excited. she had practiced late but she didn't know \textcolor{blue}{how to others}. she finally decided to try out. \\ she tried it, but glad she was it} \\
     \cmidrule{2-3}
     & + Ours (20) & \makecell[l]{she was both nervous and excited. she had a routine, but didn't know how to try. she finally decided to try out. she made \\ it and was glad she made it.} \\
     \cmidrule{2-3}
     & Gold & \makecell[l]{she started practicing every week. she was getting excited as tryouts got closer.\\ on the day of tryouts, she was ready. when the list of who made it was posted, sara's name was on it.} \\

\midrule
\multirow{1}{*}{6} & Prompt & susie wanted an ipad. \\
    \cmidrule{2-3}
     & Origin (2000) & \makecell[l]{she thought it would be great. finally she went her parents to get one. but \textcolor{blue}{when she opened it was gone}. she couldn't \\ wait to use.} \\
     \cmidrule{2-3}
     & + Ours (2000) & \makecell[l]{she thought it would be fun. finally she convinced her parents to get one. when she opened it it was great. she couldn't \\ wait to buy it.}\\
     \cmidrule{2-3}
     & + Respace (20) & \makecell[l]{she thought it would be great. finally she convinced her family to get one. when she opened it up it was all great. she \\ couldn't even use it.} \\
     \cmidrule{2-3}
     & + Ours (20) & \makecell[l]{she thought it would be great. finally she convinced her parents to get one. when she opened it on it was beautiful. \\ she couldn't wait to use it.} \\
     \cmidrule{2-3}
     & Gold & \makecell[l]{she begged for it. finally she looked under the tree. she saw one. she immediately hugged her parents.} \\

\bottomrule
\end{tabular}}
\caption{Representative generated results of the ROC testing set, where \textcolor{blue}{blue color} denotes the serious grammatical errors.}
\label{tab:case_1}
\end{table*}
\begin{table*}[t]
\centering
\resizebox{\textwidth}{!}{
\begin{tabular}{c | l | l }
\toprule
Case & Type & Text \\
\midrule
\multirow{1}{*}{1} & Prompt & the extinct volcano in the centre of edinburgh, capital city of scotland, has been known as arthur's seat for centuries. \\
    \cmidrule{2-3}
     & Origin (2000) & \makecell[l]{what is the \textcolor{blue}{capital of : scotland}} \\
     \cmidrule{2-3}
     & + Ours (2000) & \makecell[l]{what is the \textcolor{blue}{capital of : scotland}} \\
     \cmidrule{2-3}
     & + Respace (20) & \makecell[l]{what is the capital of plane?}\\
     \cmidrule{2-3}
     & + Ours (20) & \makecell[l]{what is the capital of scotland?}\\
    \cmidrule{2-3}
     & Gold  & \makecell[l]{edinburgh castle stands on arthur's seat what was arthur's seat} \\
     
\midrule
\multirow{1}{*}{2} & Prompt &  \makecell[l]{vergil in his wonderful poem, the aeneid, in two or three places, speaks of the queen of heaven, the goddess juno, as \\ descending clothed with luminous light, and he uses the word nimbus for this} \\
    \cmidrule{2-3}
     & Origin (2000) & \makecell[l]{ae was the luminous goddess \textcolor{blue}{of what}} \\
     \cmidrule{2-3}
     & + Ours (2000) & \makecell[l]{what was the luminous goddess of greek with} \\
     \cmidrule{2-3}
     & + Respace (20) &\makecell[l]{who was the luminous goddess \textcolor{blue}{of what}} \\
     \cmidrule{2-3}
     & + Ours (20) &\makecell[l]{how many was the greek goddess of light} \\
    \cmidrule{2-3}
     & Gold  & \makecell[l]{what's the word for the luminous mist that surrounds a saint} \\
     
\midrule
\multirow{1}{*}{3} & Prompt & regulation time and overtime each game consists of three periods of twenty minutes each. \\
    \cmidrule{2-3}
     & Origin (2000) & \makecell[l]{there are minutes singer, how many in a game of?} \\
     \cmidrule{2-3}
     & + Ours (2000) & \makecell[l]{how many minutes is each period of hockey} \\
     \cmidrule{2-3}
     & + Respace (20) &\makecell[l]{what is the puck arery called in a game of hockey} \\
     \cmidrule{2-3}
     & + Ours (20) &\makecell[l]{how many minutes is each period of hockey} \\
     \cmidrule{2-3}
     & Gold  & \makecell[l]{how many minutes is each period of hockey?} \\

\midrule
\multirow{1}{*}{4} & Prompt & \makecell[l]{it is then dipped into powdered borax, and the adhering solid is held in the hottest part of the flame \\ where it swells up as it loses its water of crystallization and then shrinks, forming a colourl} \\
    \cmidrule{2-3}
     & Origin (2000) & \makecell[l]{what colour is hottests} \\
     \cmidrule{2-3}
     & + Ours (2000) & \makecell[l]{what colour is hottests}\\
     \cmidrule{2-3}
     & + Respace (20) & \makecell[l]{what colour is hottest featured into a govern} \\
     \cmidrule{2-3}
     & + Ours (20) & \makecell[l]{what colour is hottests} \\
     \cmidrule{2-3}
     & Gold  & \makecell[l]{when traces of a calcium compound are held in a bunsen flame, the colour of the flame changes to \_ \_ \_?} \\

\midrule
\multirow{1}{*}{5} & Prompt & a name for the kestrel : cf. windhover. \\
    \cmidrule{2-3}
     & Origin (2000) & \makecell[l]{the windhover is the country name for which bird} \\
     \cmidrule{2-3}
     & + Ours (2000) & \makecell[l]{the windhover is an alternative name for which bird?}\\
     \cmidrule{2-3}
     & + Respace (20) & \makecell[l]{the windhover is the country name for which bird} \\
     \cmidrule{2-3}
     & + Ours (20) & \makecell[l]{the windhover is an alternative name for which bird?} \\
     \cmidrule{2-3}
     & Gold & \makecell[l]{what is the moons astronomical name} \\

\midrule
\multirow{1}{*}{6} & Prompt & to help direct the path of light made available by clerestory windows, use light colored surfaces on your interior walls. \\
    \cmidrule{2-3}
     & Origin (2000) & \makecell[l]{what is the term for the \textcolor{blue}{path path} of a body a weight} \\
     \cmidrule{2-3}
     & + Ours (2000) & \makecell[l]{what is the term for the path of a flag made in space}\\
     \cmidrule{2-3}
     & + Respace (20) & \makecell[l]{what is the term for the path followed by a body light?} \\
     \cmidrule{2-3}
     & + Ours (20) & \makecell[l]{path or trajectory of a annually, \textcolor{blue}{light by a credited}.} \\
     \cmidrule{2-3}
     & Gold  & \makecell[l]{a row of windows in the upper part of a wall, especially in a church, to admit light below.} \\

\bottomrule
\end{tabular}}
\caption{Representative generation results on Quasar-T testing set, where \textcolor{blue}{blue color} denotes the serious grammatical errors.}
\label{tab:case_1}
\end{table*}
\centering
\begin{table*}[t]
\resizebox{\textwidth}{!}{
\begin{tabular}{c | l | l }
\toprule
Case & Type & Text \\
\midrule
\multirow{1}{*}{1} & Prompt & what are some of the pros and cons of a hillary clinton presidency? \\
    \cmidrule{2-3}
     & Origin (2000) & \makecell[l]{what are some of \textcolor{blue}{pros cons} of hillary clinton presidency?} \\
     \cmidrule{2-3}
     & + Ours (2000) & \makecell[l]{what are the pros and cons of hillary clinton presidency?} \\
     \cmidrule{2-3}
     & + Respace (20) & \makecell[l]{what are pros and cons of \textbackslash \textcolor{red}{u201d} and hillary clinton?}\\
     \cmidrule{2-3}
     & + Ours (20) & \makecell[l]{what are the pros and cons of hillary clinton?}\\
     \cmidrule{2-3}
     & Gold & \makecell[l]{what are pros and cons of hillary clinton?}\\
\midrule
\multirow{1}{*}{2} & Prompt & which is less healthy for beer, your body? \\
    \cmidrule{2-3}
     & Origin (2000) & \makecell[l]{which is less healthy for your body, beer or \textcolor{red}{pepsi}?} \\
     \cmidrule{2-3}
     & + Ours (2000) & \makecell[l]{which one is better healthy, beer or \textcolor{red}{pepsi}?} \\
     \cmidrule{2-3}
     & + Respace (20) &\makecell[l]{which is less healthy \textcolor{blue}{healthy fors}, beer or body?} \\
     \cmidrule{2-3}
     & + Ours (20) &\makecell[l]{which one is better for your body, beer or pepsi?} \\
     \cmidrule{2-3}
     & Gold & \makecell[l]{which is less healthy for your body light beer or pepsi?}\\
\midrule
\multirow{1}{*}{3} & Prompt & is evolution real? how has this been proven? \\
    \cmidrule{2-3}
     & Origin (2000) & \makecell[l]{how evolution is real?} \\
     \cmidrule{2-3}
     & + Ours (2000) & \makecell[l]{how does evolution really exist?} \\
     \cmidrule{2-3}
     & + Respace (20) &\makecell[l]{is evolution really real?} \\
     \cmidrule{2-3}
     & + Ours (20) &\makecell[l]{is evolution really real?} \\
     \cmidrule{2-3}
     & Gold & \makecell[l]{what are the different ways in which evolution has been proven to be real?}\\
     
\midrule
\multirow{1}{*}{4} & Prompt & how can i impress a girl who hate me? \\
    \cmidrule{2-3}
     & Origin (2000) & \makecell[l]{how can i impress a girl \textcolor{red}{$[\cdots]$}?} \\
     \cmidrule{2-3}
     & + Ours (2000) & \makecell[l]{what should i do to impress a girl hate me?}\\
     \cmidrule{2-3}
     & + Respace (20) & \makecell[l]{how do i impress a girl \textcolor{red}{$[\cdots]$}}?  \\
     \cmidrule{2-3}
     & + Ours (20) & \makecell[l]{how do i impress a girl who hate me?} \\
     \cmidrule{2-3}
     & Gold & \makecell[l]{how can impress a girl who hate me?}\\

\midrule
\multirow{1}{*}{5} & Prompt & what are the best and most effective openings in chess? \\
    \cmidrule{2-3}
     & Origin (2000) & \makecell[l]{what are the best effective \textcolor{red}{\textbackslash u201d} in chess?} \\
     \cmidrule{2-3}
     & + Ours (2000) & \makecell[l]{what are the best effective openings in chess?}\\
     \cmidrule{2-3}
     & + Respace (20) & \makecell[l]{what are the best effective \textcolor{red}{$[\cdots]$} in chess?} \\
     \cmidrule{2-3}
     & + Ours (20) & \makecell[l]{what are the best effective openings in chess?} \\
     \cmidrule{2-3}
     & Gold & \makecell[l]{what are the best opening moves in chess?}\\

\midrule
\multirow{1}{*}{6} & Prompt & if you were given a chance to be somebody else for a day, who would you be? \\
    \cmidrule{2-3}
     & Origin (2000) & \makecell[l]{if you were a chance \textcolor{red}{for his} a day, who would you be?} \\
     \cmidrule{2-3}
     & + Ours (2000) & \makecell[l]{if you were given a chance to be somebody else for a day, who would you be?}\\
     \cmidrule{2-3}
     & + Respace (20) & \makecell[l]{if you were given in a day \textcolor{red}{who will be else}, who would it be?} \\
     \cmidrule{2-3}
     & + Ours (20) & \makecell[l]{if you were given a chance to be else for a day, who would you be?} \\
     \cmidrule{2-3}
     & Gold & \makecell[l]{if you were given a chance to be somebody else for one day, who would you be?}\\

\midrule
\multirow{1}{*}{7} & Prompt & i want a real and effective way to make \$ 500 per month with the knowledge that i have no money to invest? \\
    \cmidrule{2-3}
     & Origin (2000) & \makecell[l]{i want a real effective way to make 500 and \textcolor{red}{\$ knowledge} per month?} \\
     \cmidrule{2-3}
     & + Ours (2000) & \makecell[l]{what is the best way to make \$ 500 fast in india?}\\
     \cmidrule{2-3}
     & + Respace (20) & \makecell[l]{i want \textcolor{blue}{a real and 500 way} to make \textbackslash u201d her month?} \\
     \cmidrule{2-3}
     & + Ours (20) & \makecell[l]{i want a real way to make \$ 500 month with the knowledge?} \\
     \cmidrule{2-3}
     & Gold & \makecell[l]{the real effective way to make \$ 500 per month with the knowledge that i have no money to invest?}\\


\bottomrule
\end{tabular}}
\caption{Representative generated results of the QQP testing set, where \textcolor{red}{red color} denotes the phrase contradicts the prompt, and \textcolor{blue}{blue color} denotes the serious grammatical errors.}
\label{tab:case_qg}
\end{table*}
\centering
\begin{table*}[t]
\resizebox{\textwidth}{!}{
\begin{tabular}{c | l | l }
\toprule
Case & Type & Text \\
\midrule
\multirow{1}{*}{1} & Prompt & construction of changdeok palace began in 1405, and was completed in 1412. \\
    \cmidrule{2-3}
     & Origin (2000) & \makecell[l]{it was completed in 1412.} \\
     \cmidrule{2-3}
     & + Ours (2000) & \makecell[l]{construction of changdeok palace began in 1405, and was completed in 1412.} \\
     \cmidrule{2-3}
     & + Respace (20) & \makecell[l]{construction of changdeok palace began in 1405.}\\
     \cmidrule{2-3}
     & + Ours (20) & \makecell[l]{construction of changdeok palace began in 1405, and was completed in 1412.}\\
     \cmidrule{2-3}
     & Gold (20) & \makecell[l]{The construction began by King Taejong, the Third King of the Joseon Dynasty, in 1405.}\\

\midrule
\multirow{1}{*}{2} & Prompt & however, this feature is not present in all cottontails nor is it unique to the genus. \\
    \cmidrule{2-3}
     & Origin (2000) & \makecell[l]{however, this feature is not present in all cottontails nor is it unique to the genus.} \\
     \cmidrule{2-3}
     & + Ours (2000) & \makecell[l]{however, this feature is not present in all cottontails nor is it unique to the genus.} \\
     \cmidrule{2-3}
     & + Respace (20) &\makecell[l]{however, this feature is not present in all cottontails nor is it unique to the genus.} \\
     \cmidrule{2-3}
     & + Ours (20) &\makecell[l]{however, this feature is not present in all cottontails nor is it unique to the genus.} \\
     \cmidrule{2-3}
     & Gold (20) & \makecell[l]{However, this feature is not present in all cottontails.}\\
     
\midrule
\multirow{1}{*}{3} & Prompt & the team was owned by ralph wilson from the team's founding in 1960, until his death in 2014 at the age of 95. \\
    \cmidrule{2-3}
     & Origin (2000) & \makecell[l]{the team was owned by ralph wilson from the team's founding in 1960.} \\
     \cmidrule{2-3}
     & + Ours (2000) & \makecell[l]{the team was owned by ralph wilson from the team's founding in 1960, until his death in 2014 at the age of 95. }\\
     \cmidrule{2-3}
     & + Respace (20) & \makecell[l]{the team was owned by ralph wilson from the team's founding in 1960.} \\
     \cmidrule{2-3}
     & + Ours (20) & \makecell[l]{the team was owned by ralph wilson from the team's founding in 1960.} \\
     \cmidrule{2-3}
     & Gold (20) & \makecell[l]{Ralph Wilson, the longtime owner who had established the Bills in 1959, died on March 25, 2014.}\\

\midrule
\multirow{2}{*}{4} & Prompt & \makecell[l]{the association first convened in the 1960s, as interest increased in the new science of modern linguistics and \\ particularly in its practical application - for example, in language teaching and learning.} \\
    \cmidrule{2-3}
     & Origin (2000) & \makecell[l]{the association first made in the 1960s.} \\
     \cmidrule{2-3}
     & + Ours (2000) & \makecell[l]{the association first convened in the 1960s, as interest increased in the new science of modern linguistics\\ and particularly in its practical application - for example, in language teaching and learning.}\\
     \cmidrule{2-3}
     & + Respace (20) & \makecell[l]{the association first made in the 1960s as interest increased in the new science of modern linguistics.} \\
     \cmidrule{2-3}
     & + Ours (20) & \makecell[l]{the association first convened in the 1960s, as interest increased in the new science of modern linguistics\\ and particularly in its practical application - for example, in language teaching and learning.} \\
     \cmidrule{2-3}
     & Gold (20) & \makecell[l]{The association started in the 1960s.}\\

\midrule
\multirow{1}{*}{5} & Prompt & his hair shorn and now blind and shackled, samson is turning a mill - wheel and praying for his people, who will suffer for his sin. \\
    \cmidrule{2-3}
     & Origin (2000) & \makecell[l]{samson is turning a mill - \textcolor{blue}{wheel and who} will suffer for his sin.} \\
     \cmidrule{2-3}
     & + Ours (2000) & \makecell[l]{his hair shorn and now blind and shackled, samson is turning a mill - wheel and praying for his people, who will suffer for his sin.}\\
     \cmidrule{2-3}
     & + Respace (20) & \makecell[l]{he is a mill - \textcolor{blue}{wheel for praying}, who will suffer for his sin.} \\
     \cmidrule{2-3}
     & + Ours (20) & \makecell[l]{his hair shorn and now blind and shackled, samson is turning a mill - wheel and praying for his people, who will suffer for his sin.} \\
     \cmidrule{2-3}
     & Gold (20) & \makecell[l]{He prays for his people.}\\

\midrule
\multirow{1}{*}{6} & Prompt & \makecell[l]{he was a significant figure in the development of ballroom dance\\ during the first half of the 20th century, and his records sold 75 million copies from the 1930s through to the 1980s.} \\
    \cmidrule{2-3}
     & Origin (2000) & \makecell[l]{his records sold 75 million copies from the 1930s through to the 1980s.} \\
     \cmidrule{2-3}
     & + Ours (2000) & \makecell[l]{he was a significant figure in the development of ballroom dance during\\ the first half of the 20th century, and his records sold 75 million copies from the 1930s through to the 1980s.}\\
     \cmidrule{2-3}
     & + Respace (20) & \makecell[l]{he was a significant figure in the development of ballroom dance during the 20th century.} \\
     \cmidrule{2-3}
     & + Ours (20) & \makecell[l]{he was a significant figure in the development of ballroom dance during the first half of the 20th century.} \\
     \cmidrule{2-3}
     & Gold (20) & \makecell[l]{He was a significant figure in the development of ballroom dance during the first half of the 20th century.}\\

\midrule
\multirow{1}{*}{7} & Prompt & alyosha monument, murmansk or defenders of the soviet arctic during the great patriotic war monument is also located in murmansk. \\
    \cmidrule{2-3}
     & Origin (2000) & \makecell[l]{alyosha monument, murmansk or defenders of the soviet arctic during the great patriotic war monument.} \\
     \cmidrule{2-3}
     & + Ours (2000) & \makecell[l]{alyosha monument, murmansk or defenders of the soviet arctic during the great patriotic war monument is also located in murmansk. }\\
     \cmidrule{2-3}
     & + Respace (20) & \makecell[l]{alyosha monument, murmansk or defenders of the soviet arctic during the great patriotic war monument.} \\
     \cmidrule{2-3}
     & + Ours (20) & \makecell[l]{alyosha monument, murmansk or defenders of the soviet arctic during the great patriotic war monument is also located\textcolor{red}{$[\cdots]$}.} \\
     \cmidrule{2-3}
     & Gold (20) & \makecell[l]{It is called the Alyosha Monument.}\\

\midrule
\multirow{1}{*}{8} & Prompt & \makecell[l]{singaporean citizens, government and non - governmental organisations may display or fly the national flag throughout the year to\\ identify themselves with the nation, and especially encouraged to do so during occasions of national celebration or national significance.} \\
    \cmidrule{2-3}
     & Origin (2000) & \makecell[l]{singaporean citizens, government and non - governmental organisations may\\ display or fly the national flag throughout the year to identify themselves with the nation.} \\
     \cmidrule{2-3}
     & + Ours (2000) & \makecell[l]{singaporean citizens, government and non - governmental organisations may display or fly the national flag throughout the year to\\ identify themselves with the nation, and especially encouraged to do so during occasions of national celebration or national significance.}\\
     \cmidrule{2-3}
     & + Respace (20) & \makecell[l]{singaporean citizens, government and non - governmental organisations may\\ display or fly the national flag throughout the year to identify themselves with the nation.} \\
     \cmidrule{2-3}
     & + Ours (20) & \makecell[l]{singaporean citizens, government and non - governmental organisations may\\ display or fly the national flag throughout the year to identify themselves with the nation.} \\
     \cmidrule{2-3}
     & Gold (20) & \makecell[l]{Singaporeans are encouraged to do this during occasions of national celebration or national significance.}\\

\bottomrule
\end{tabular}}
\caption{Representative generated results of the WIKI-AUTO testing set, where \textcolor{red}{red color} denotes the phrase contradicts the prompt, and \textcolor{blue}{blue color} denotes the serious grammatical errors.
We can observe that our method can generate much shorter and streamlined content.}
\label{tab:case_paraphrase}
\end{table*}
\begin{table*}[t]
\centering
\small
\resizebox{\textwidth}{!}{
\begin{tabular}{c | l | l }
\toprule
Case & Type & Text \\
\midrule
\multirow{1}{*}{1} & Prompt & name : The Vaults \\
    \cmidrule{2-3}
     & Origin (200) & \makecell[l]{\textcolor{red}{The Vaults Two} is a family friendly Italian restaurant .} \\
     \cmidrule{2-3}
     & + Ours (200) & \makecell[l]{\textcolor{green}{The Vaults} is a cheap , family friendly Italian restaurant .} \\
     \cmidrule{2-3}
     & + Respace (20) & \makecell[l]{\textcolor{red}{The Plough} is a cheap , family friendly pub near Caf\textbackslash u00e9 Rouge .}\\
     \cmidrule{2-3}
     & + Ours (20) & \makecell[l]{\textcolor{green}{The Vaults} is a family friendly fast food restaurant .}\\
     
\midrule
\multirow{1}{*}{2} & Prompt & name : The Cambridge Blue \\
    \cmidrule{2-3}
     & Origin (200) & \makecell[l]{\textcolor{green}{The Cambridge Blue} provides Indian food \textcolor{blue}{Its} customer rating is low .} \\
     \cmidrule{2-3}
     & + Ours (200) & \makecell[l]{\textcolor{green}{The Cambridge Blue} is a restaurant that serves Italian food .} \\
     \cmidrule{2-3}
     & + Respace (20) &\makecell[l]{\textcolor{red}{Browns Cambridge} is a 5 star dine in restaurant . It is moderately priced .} \\
     \cmidrule{2-3}
     & + Ours (20) &\makecell[l]{\textcolor{green}{The Cambridge Blue} is a restaurant with a high customer rating .} \\

\midrule
\multirow{1}{*}{3} & Prompt & name : The Golden Palace \\
    \cmidrule{2-3}
     & Origin (200) & \makecell[l]{\textcolor{red}{The Mill} is a coffee shop that serves Italian food, is located in riverside near The Sorrento .} \\
     \cmidrule{2-3}
     & + Ours (200) & \makecell[l]{\textcolor{green}{The Golden Palace} is a high priced coffee shop serving Indian food located in the city centre with a customer rating of 1 out of 5 .} \\
     \cmidrule{2-3}
     & + Respace (20) &\makecell[l]{\textcolor{green}{The Golden Palace} is a Japanese coffee shop with a moderate price range in the city centre . \textcolor{blue}{1 out of 5 customer rating} .} \\
     \cmidrule{2-3}
     & + Ours (20) &\makecell[l]{\textcolor{green}{The Golden Palace} is a fast food coffee shop in the city centre that has a moderate price range and a customer rating of 1 out of 5 .} \\

\midrule
\multirow{1}{*}{4} & Prompt & Type : pub \\
    \cmidrule{2-3}
     & Origin (200) & \makecell[l]{ Blue Spice provides Chinese food in the high price range .\\ It is located in the \textcolor{red}{city centre} .} \\
     \cmidrule{2-3}
     & + Ours (200) & \makecell[l]{The Olive Grove is a \textcolor{green}{pub} providing Chinese food It is located in the city centre .}\\
     \cmidrule{2-3}
     & + Respace (20) & \makecell[l]{Wildwood , a \textcolor{green}{pub} \textcolor{blue}{serving} French food with a customer rating of low .} \\
     \cmidrule{2-3}
     & + Ours (20) & \makecell[l]{The Mill is a \textcolor{green}{pub} that provides Indian food It is in the cheap price range .\\ It is located in riverside .} \\

\midrule
\multirow{1}{*}{5} & Prompt & near : Clare Hall \\
    \cmidrule{2-3}
     & Origin (200) & \makecell[l]{There is a cheap family friendly Japanese restaurant called Loch Fyne . \textbackslash n  END near \textcolor{red}{Clare}} \\
     \cmidrule{2-3}
     & + Ours (200) & \makecell[l]{Bibimbap House is a Chinese restaurant in the high price range .  It is located in the riverside area near \textcolor{green}{Clare Hall} .}\\
     \cmidrule{2-3}
     & + Respace (20) & \makecell[l]{This restaurant \textcolor{blue}{Bibimbap House Clare Hall} is a cheap and located in the city that serves Japanese food .} \\
     \cmidrule{2-3}
     & + Ours (20) & \makecell[l]{Clowns is a coffee shop by the riverside near \textcolor{green}{Clare Hall} and has a customer rating of 3 out of 5 .} \\

\midrule
\multirow{1}{*}{6} & Prompt & near : The Six Bells \\
    \cmidrule{2-3}
     & Origin (200) & \makecell[l]{Near \textcolor{green}{The Six Bells} is \textcolor{blue}{Fitzbillies , , cheap ,} which serves English food .} \\
     \cmidrule{2-3}
     & + Ours (200) & \makecell[l]{Fitzbillies is a moderately priced Italian restaurant located near \textcolor{green}{The Six Bells} .}\\
     \cmidrule{2-3}
     & + Respace (20) & \makecell[l]{Near \textcolor{green}{The Six Bells} , \textcolor{blue}{Giraffe a moderately priced restaurant .}} \\
     \cmidrule{2-3}
     & + Ours (20) & \makecell[l]{Giraffe is a restaurant near \textcolor{green}{The Six Bells} with a high price range .} \\

\midrule
\multirow{1}{*}{7} & Prompt & family friendly : yes \\
    \cmidrule{2-3}
     & Origin (200) & \makecell[l]{The Eagle is a fast food restaurant that is \textcolor{red}{highly rated}} \\
     \cmidrule{2-3}
     & + Ours (200) & \makecell[l]{Loch Fyne is a \textcolor{green}{family friendly} restaurant that serves English food .}\\
     \cmidrule{2-3}
     & + Respace (20) & \makecell[l]{Zizzi is \textcolor{blue}{a , 1} star restaurant , it offers spirits , and it is \textcolor{green}{family friendly}} \\
     \cmidrule{2-3}
     & + Ours (20) & \makecell[l]{The Cricketers is a \textcolor{green}{family friendly} coffee shop serving Japanese food .\\ It is located near The Portland Arms with a customer rating of 5 out of 5 .} \\

\midrule
\multirow{1}{*}{8} & Prompt & food : Chinese \\
    \cmidrule{2-3}
     & Origin (200) & \makecell[l]{The Wrestlers provides \textcolor{green}{Chinese food} in the moderate price range \textcolor{blue}{.customer rating} is 1 out of 5 .} \\
     \cmidrule{2-3}
     & + Ours (200) & \makecell[l]{The Waterman is providing \textcolor{green}{Chinese food} in the cheap price range . It is located in the riverside . Its customer rating is \\ 5 out of 5 .}\\
     \cmidrule{2-3}
     & + Respace (20) & \makecell[l]{Browns Cambridge sells \textcolor{red}{Japanese food} , for 20 - 25 , with a customer rating 3 out of 5} \\
     \cmidrule{2-3}
     & + Ours (20) & \makecell[l]{The Waterman provides \textcolor{green}{Chinese food} in the cheap price range . It is located in the riverside . Its customer rating is average .} \\


\bottomrule
\end{tabular}}
\caption{Representative generated results of the E2E~(Semantic Content) testing set, where \textcolor{green}{green color} denotes the key phrase related to the prompt, \textcolor{red}{red color} denotes the phrase contradicts the prompt, and \textcolor{blue}{blue color} denotes the serious grammatical errors.}
\label{tab:case_semantic_control}
\end{table*}
\begin{table*}[t]
\centering
\resizebox{\textwidth}{!}{
\begin{tabular}{c | l | l }
\toprule
Case & Type & Text \\
\midrule
\multirow{1}{*}{1} & Prompt & [3, 5, PP] \\
    \cmidrule{2-3}
     & Origin (200) & \makecell[l]{Browns Cambridge is \textcolor{green}{in the city} centre near The Sorrento . It is a family - friendly restaurant .} \\
     \cmidrule{2-3}
     & + Ours (200) & \makecell[l]{In city centre \textcolor{green}{near Clare Hall} , there is a coffee shop called Clowns . It also serves Italian food and has a customer rating of 5 out of 5 .} \\
     \cmidrule{2-3}
     & + Respace (20) & \makecell[l]{Browns Cambridge is \textcolor{green}{on the riverside} near The Sorrento . It is a family friendly restaurant serving English food .}\\
     \cmidrule{2-3}
     & + Ours (20) & \makecell[l]{Aromi is located \textcolor{green}{in the city} centre and is a family - friendly coffee shop that serves Italian food with a low customer rating .}\\
     \cmidrule{2-3}
     & Gold & \makecell[l]{Only feet away from Caf\u00e9 Sicilia , The Punter coffee Shop offers low price coffee and does not have family restrooms .}\\
\midrule
\multirow{1}{*}{2} & Prompt & [5, 7, NP] \\
    \cmidrule{2-3}
     & Origin (200) & \makecell[l]{Cotto provides Indian food It \textcolor{red}{is near Ranch} . Its customer rating is average .} \\
     \cmidrule{2-3}
     & + Ours (200) & \makecell[l]{In the city centre near \textcolor{green}{The Portland Arms} , there is a coffee shop called Cotto . It serves Italian food . It has a high price range and a \\ customer rating of 1 out of 5 .} \\
     \cmidrule{2-3}
     & + Respace (20) &\makecell[l]{The Vaults is a restaurant \textcolor{red}{providing Italian food} in the high price range .} \\
     \cmidrule{2-3}
     & + Ours (20) &\makecell[l]{In the city centre near \textcolor{green}{Crowne Plaza Hotel} is a fast food coffee shop named Browns Cambridge . It has a customer rating of 5 out of 5\\ and is family - friendly .} \\
     \cmidrule{2-3}
     & Gold & \makecell[l]{For date night go to The Rice Boat , cheap , average rated Chinese not family friendly food near Express by Holiday Inn}\\
\midrule
\multirow{1}{*}{3} & Prompt & [6, 9, ADVP] \\
    \cmidrule{2-3}
     & Origin (200) & \makecell[l]{Blue Spice provides Chinese food in \textcolor{red}{the high price range} . It is located in the city centre .} \\
     \cmidrule{2-3}
     & + Ours (200) & \makecell[l]{Cotto is a cheap restaurant located \textcolor{green}{near All Bar One} } \\
     \cmidrule{2-3}
     & + Respace (20) &\makecell[l]{The Plough is a cheap Italian \textcolor{red}{pub near Caf\textbackslash u00e9 Rouge} . It is family friendly .} \\
     \cmidrule{2-3}
     & + Ours (20) &\makecell[l]{Clowns is a coffee shop located \textcolor{green}{next to Clare Hall} .} \\
     \cmidrule{2-3}
     & Gold & \makecell[l]{Browns Cambridge located in city centre close to The Sorrento serves Indian food . It is a adult dining restaurant .}\\
\midrule
\multirow{1}{*}{4} & Prompt & [10, 11, PP] \\
    \cmidrule{2-3}
     & Origin (200) & \makecell[l]{Aromi is a coffee shop that is rated 5 out \textcolor{red}{of 5} and serves French food . It is not family - friendly . It is located in the city centre .} \\
     \cmidrule{2-3}
     & + Ours (200) & \makecell[l]{The Rice Boat is located near Express by Holiday Inn \textcolor{green}{in riverside} . It is a family - friendly Japanese restaurant \\with a low customer rating and a low price range .}\\
     \cmidrule{2-3}
     & + Respace (20) & \makecell[l]{The Twenty Two offers Japanese food in a family - \textcolor{red}{friendly environment} . It is located in the city centre .} \\
     \cmidrule{2-3}
     & + Ours (20) & \makecell[l]{The Eagle coffee shop has a rating of 5 out \textcolor{red}{of 5} . It is a family friendly Fast food place in the city centre , near Burger King .} \\
     \cmidrule{2-3}
     & Gold & \makecell[l]{Highly rated English food restaurant The Waterman , is located in Riverside . The cost is high but is not child friendly .}\\

\midrule
\multirow{1}{*}{5} & Prompt & [0, 0, NP] \\
    \cmidrule{2-3}
     & Origin (200) & \makecell[l]{\textcolor{red}{There} is a family friendly Japanese restaurant in the riverside area near The Sorrento , named Browns Cambridge .} \\
     \cmidrule{2-3}
     & + Ours (200) & \makecell[l]{\textcolor{green}{Fitzbillies} is a coffee shop providing Indian food in the high price range . It is located in the city centre . Its customer rating is 1 out of 5 .}\\
     \cmidrule{2-3}
     & + Respace (20) & \makecell[l]{\textcolor{red}{There} is a kid - friendly fast food restaurant in Riverside called The Twenty Two .} \\
     \cmidrule{2-3}
     & + Ours (20) & \makecell[l]{\textcolor{green}{Wildwood} is a coffee shop providing Indian food in the moderate price range . It is near Ranch . Its customer rating is 1 out of 5 .} \\
     \cmidrule{2-3}
     & Gold & \makecell[l]{There is a family friendly coffee shop located close to the Crowne Plaza Hotel . It is called Browns Cambridge .}\\

\midrule
\multirow{1}{*}{6} & Prompt & [4, 6, NP] \\
    \cmidrule{2-3}
     & Origin (200) & \makecell[l]{Taste of Cambridge is \textcolor{green}{a coffee shop} that serves Japanese food . It is located in the riverside area near Crowne Plaza Hotel . It is not \\ family - friendly .} \\
     \cmidrule{2-3}
     & + Ours (200) & \makecell[l]{The Eagle is \textcolor{blue}{a} \textcolor{green}{a coffee shop} that provides Indian food in the high price range . It is located in the riverside . It is near Burger King . \\Its customer rating is 1 out of 5 .}\\
     \cmidrule{2-3}
     & + Respace (20) & \makecell[l]{Alimentum is located in \textcolor{green}{the city centre} and serves Japanese food . It is kid friendly and has a price range of \u00a3 20 - 25 .} \\
     \cmidrule{2-3}
     & + Ours (20) & \makecell[l]{The Rice Boat is \textcolor{green}{a Japanese restaurant} located in the city centre near the Express by Holiday Inn . It is kid friendly and has a price range \\ of 20 - 25 . It has a customer rating of 3 out of 5 .} \\
     \cmidrule{2-3}
     & Gold & \makecell[l]{The Golden Palace is a coffee shop with Italian food , prices less then 20 , in the riverside and has low ratings .}\\

\bottomrule
\end{tabular}}
\caption{Representative generated results of the E2E~(Syntax Spans) testing set, where \textcolor{green}{green color} denotes the key phrase related to the prompt, \textcolor{red}{red color} denotes the phrase contradicts the prompt, and \textcolor{blue}{blue color} denotes the serious grammatical errors.}
\label{tab:case_syntax_control}
\end{table*}

\end{document}